\documentclass{article}

\PassOptionsToPackage{numbers,sort&compress}{natbib}
\usepackage[eandd,final]{neurips_2026}

\usepackage[utf8]{inputenc}
\usepackage[T1]{fontenc}
\usepackage{hyperref}
\hypersetup{colorlinks=true,linkcolor=[rgb]{0.10,0.25,0.55},citecolor=[rgb]{0.05,0.40,0.25},urlcolor=[rgb]{0.10,0.25,0.55},
  pdftitle={WirelessMathBench-XL: An Auditable Benchmark for Wireless Mathematical Reasoning},
  pdfauthor={Xin Li, Mengbing Liu, Yiyang Zhu, Wenhe Zhang, Li Wei, Jiancheng An, Chau Yuen}}
\usepackage{url}
\usepackage{booktabs}
\usepackage{array}
\usepackage{amsfonts}
\usepackage{amsmath, amssymb}
\usepackage{nicefrac}
\usepackage{microtype}
\usepackage{xcolor}
\usepackage{graphicx}
\usepackage[most]{tcolorbox}
\usepackage{needspace}

\definecolor{correctgreen}{HTML}{2E8B57}
\newtcolorbox{questionbox}[1][]{colback=blue!2,colframe=blue!40,
  title={#1},breakable,fonttitle=\bfseries\small}
\newtcolorbox{equationbox}{colback=gray!5,colframe=gray!40,breakable}

\graphicspath{{figures/}}

\title{WirelessMathBench-XL: An Auditable Benchmark for
Wireless Mathematical Reasoning}

\author{%
  Xin Li \quad Mengbing Liu \quad Yiyang Zhu \quad Wenhe Zhang \\
  \textbf{Li Wei} \quad \textbf{Jiancheng An} \quad \textbf{Chau Yuen} \\
  Nanyang Technological University \\
  \href{https://lixin.ai/WirelessMathBench-XL}{\texttt{lixin.ai/WirelessMathBench-XL}}
}

\begin{document}
\maketitle

\begin{abstract}
Benchmarks built from arXiv papers can overlap the arXiv text used in LLM
pretraining, yet technical-domain releases rarely ship the per-item evidence
needed to check this. We introduce a \textbf{reverse-probe $n$-gram audit}
that indexes the benchmark side and streams the corpus past it, so memory
scales with the benchmark rather than the corpus and flags on the distributed
problems can be inspected and recomputed. We instantiate it in
\textbf{WirelessMathBench-XL}, a $4{,}027$-problem benchmark for wireless
mathematical reasoning built from $836$ arXiv papers across $20$ subfields,
with multiple-choice, progressive fill-in, and full-equation-completion
formats. Streaming $12.9$\,B RedPajama-arXiv $13$-grams yields a strict
zero-hit view $\mathcal{S}_0$ covering $95.7\%$ of the problems, and shows
that the Dolma and ProofPile-2 arXiv slices repackage the same RedPajama dump.
Filtering to $\mathcal{S}_0$ moves every evaluated model by less than
$1$\,pp. With $30/800$ flagged test items, this comparison bounds
detected-overlap inflation at $0.31$--$0.51$\,pp for frontier models rather
than testing for contamination, and it does not cover paraphrase,
target-answer, alternate-version, post-training, or closed-corpus exposure.
Frontier models reach $86.5$--$91.3\%$, and GRPO on the released training
split lifts a $7$B model $25.1$\,pp above a size-matched general-math
control. We release source-paper identifiers, per-problem audit and
threshold metadata, license-aware data files, audit and bootstrap scripts,
Croissant metadata, and a Datasheet.
\end{abstract}

\section{Introduction}
\label{sec:intro}

When a frontier model solves a wireless-math benchmark item, it may be
deriving the equation from the stated constraints, or it may have seen the
source-paper text during pretraining. This ambiguity is especially sharp
for benchmarks built from arXiv papers: public pretraining resources such as
RedPajama, Dolma, and ProofPile-2 include arXiv-derived slices
\citep{weber2024redpajama,soldaini2024dolma,azerbayev2024llemma}, and the
relevant exposure varies by model cutoff and training mixture. Memorisation
can scale with model size and duplicated data \citep{carlini2023quantifying},
and benchmark contamination must be measured at the benchmark level
\citep{sainz2023nlp,deng2024investigating,xu2024contamsurvey}. Yet released
technical-domain benchmarks rarely ship the per-item provenance and overlap
evidence that would let users check this
\citep{gebru2021datasheets,akhtar2024croissant}.

Producing that evidence means searching billions of corpus $n$-grams while
keeping the result at the level of individual problems. Corpus-side search
systems index the corpus \citep{elazar2024wimbd}; we invert the direction.
Our \emph{reverse-probe} audit indexes the benchmark prompts, streams the
corpus past them, and accumulates per-problem hit counts. Memory therefore
scales with the benchmark. Any user can rerun the audit as corpora change,
and so can any corpus holder with a private mixture.

We instantiate the audit on wireless mathematical reasoning, where answers
hinge on implicit constraints: feasible power must remain non-negative,
complex matrices must preserve dimensions and conjugate structure,
logarithms require valid SINR domains, and causal filters cannot depend on
future channel states. Existing telecom LLM resources mostly evaluate QA or
assistant behaviour
\citep{maatouk2023teleqna,shao2024wirelessllm,zou2024telecomgpt,maatouk2024tele},
and the closest wireless-math benchmark is built from $40$ papers
\citep{li2025wirelessmathbench}. \textbf{WirelessMathBench-XL} contains
$4{,}027$ problems from $836$ arXiv papers across $20$ subfields, in three
formats that probe these constraints at increasing strictness. It comes with
source-paper identifiers, verifier-facing ground truths, a problem-level
train/test split, and two-tier QA that includes expert screening.

Against $12.9$\,B RedPajama-arXiv $13$-grams, the audit labels $95.7\%$ of
the problems zero-hit ($\mathcal{S}_0$) and finds that the Dolma and
ProofPile-2 arXiv slices are repackaged copies of the same dump, so they add
no independent in-domain evidence. Filtering to $\mathcal{S}_0$ moves every
evaluated model by less than $1$\,pp. Because only $30$ of $800$ test items
are flagged, this bounds how much detected overlap can inflate scores
through one channel, exact overlap with the prompt surface
(Section~\ref{sec:contamination-audit}), rather than certifying the
benchmark clean. On the benchmark itself, frontier models
reach $86.5$--$91.3\%$ while open $7$B models reach $40$--$48\%$, and GRPO on
the released training split lifts a $7$B model $25.1$\,pp above a
size-matched general-math control.

\noindent\textbf{Contribution 1: a reverse-probe audit with released
per-problem evidence.} The audit indexes the benchmark side, so a
$4{,}027$-problem benchmark is checked against $12.9$\,B corpus grams on one
workstation. The release ships per-problem RedPajama-arXiv hit counts,
strict and lenient cleaned-subset flags, memberships at every tested
threshold, OpenWebMath corroboration, an off-domain negative control, and
paired-bootstrap scripts, so users can tighten thresholds, audit new
corpora, or rebuild the cleaned-subset evaluation without regenerating the
benchmark.

\noindent\textbf{Contribution 2: a provenance-complete benchmark for
wireless mathematical reasoning.} WirelessMathBench-XL is $7\times$ larger
than its predecessor ($4{,}027$ vs.\ $587$ problems) and $21\times$ broader
in sources ($836$ vs.\ $40$ papers). Distractors encode dimension, sign,
domain, and causality violations; Fill-in-the-blank (Fill-in) tasks form a
$25/50/75\%$ masking ladder; and Full Equation Completion (FEC) removes
every equation component while keeping assumptions and variables. Every
record carries its \texttt{paper\_id}, and the release materialises a
source-paper-disjoint view and a grouped split for clustered evaluation.

\noindent\textbf{Contribution 3: reference evaluations and learnability
checks.} We evaluate $19$ calibration rows under one published verifier
stack and train WirelessMathLM reference checkpoints with off-the-shelf GRPO
and binary verification rewards. WirelessMathLM-$7$B ($47.88\%$) exceeds
DeepSeekMath-$7$B-RL ($41.00\%$) and Qwen$2.5$-Math-$7$B-Instruct
($40.00\%$), and the released training split outperforms a size-matched
Math12K control by $25.13$\,pp.

\section{Constructing WirelessMathBench-XL}
\label{sec:dataset_construction}

\begin{figure}[t]
\centering
\includegraphics[width=.86\linewidth]{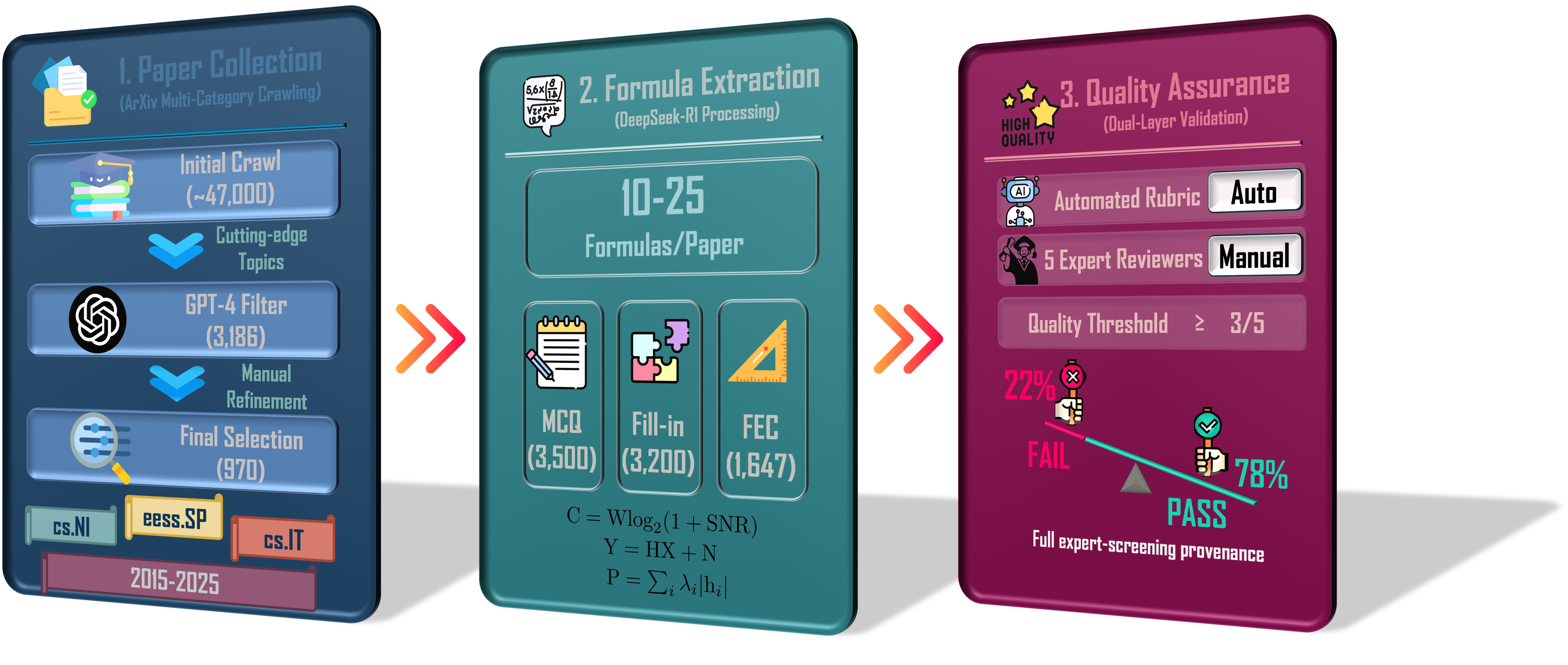}
\caption{\textbf{WirelessMathBench-XL construction pipeline.}
Papers are filtered, mined for structured mathematical records, turned
into three task formats, and screened by automated and expert QA; format
counts are pre-QA candidates, and $4{,}027$ problems are released.}
\label{fig:overview}
\end{figure}

WirelessMathBench-XL is built in four stages (Figure~\ref{fig:overview}): source collection
(Section~\ref{sec:collection}), extraction and three-format generation
(Section~\ref{sec:extraction}), two-tier QA (Section~\ref{sec:qa}), and
release statistics with the train/test split (Section~\ref{sec:stats}).
Section~\ref{sec:contamination-audit} then attaches the contamination-audit
metadata used by the cleaned-subset evaluation.

\subsection{Source Paper Collection}
\label{sec:collection}

Prior wireless-math resources use smaller source pools; for example,
WirelessMathBench~\citep{li2025wirelessmathbench} draws on $40$ papers.
WirelessMathBench-XL targets broader coverage of contemporary wireless
subfields while keeping filtering strict enough to retain mathematical
content and source-paper provenance.

We query $24$ arXiv categories spanning core wireless domains
(\texttt{cs.NI}, \texttt{eess.SP}, \texttt{cs.IT}), AI/ML
(\texttt{cs.LG}, \texttt{stat.ML}), and adjacent areas, retrieving
candidate papers from $2005$--$2025$. Each candidate
receives a deterministic relevance score in $[0,1]$ from weighted keyword
hits and arXiv-category bonuses (weights in
Appendix~\ref{app:data_collection}), with no language-model judgment.
A second-stage GPT-4o verifier then
identifies candidates that contain substantial mathematical
content. A third stage then ranks the candidates on three recorded
selection criteria: mathematical rigour, citation signal, and topical
diversity. This yields a $970$-paper source pool; after extraction and QA, $836$ retained papers
contribute accepted problems.

\subsection{Mathematical Content Extraction and Problem Generation}
\label{sec:extraction}

We extract variable definitions, channel/transmitter assumptions, and
unstated domain restrictions as explicit fields, so that the
verifier (Section~\ref{sec:qa}) can grade reasoning that depends on
conditions that the problem text never spelled out.

\noindent\textbf{Structured Model Extraction.}
We use DeepSeek-R1~\citep{guo2025deepseek} over each paper's LaTeX source to
produce a structured record per paper. Each record contains the system
equations, variable definitions with units and domain restrictions,
underlying assumptions, and boundary conditions, with notation normalised
at extraction time (e.g., $\boldsymbol{v}$ for vectors,
$\mathbf{H} \in \mathbb{C}^{N \times M}$ for complex matrices). Three
representative records (SIM-based air-ground, UAV-assisted MEC,
RIS-aided random access) are in Appendix~\ref{app:model_examples}.

\noindent\textbf{Three Task Formats.}
From each structured record we generate problems in three formats with
deliberately different difficulty profiles. \textbf{MCQ} masks the right-hand
side of a key equation and offers four options whose distractors are
\emph{constraint-violating} by construction (matrix-dimension mismatches,
operator-order errors, sign violations of implicit non-negativity), so the
subtest discriminates failures on implicit constraints rather than surface
plausibility. \textbf{Progressive Fill-in-the-blank} masks $25/50/75\%$ of
components in three rungs, exposing a difficulty gradient inside one
problem family. \textbf{Full Equation Completion (FEC)} masks all
components and supplies only variable definitions, assumptions, and
boundary conditions, making it the strictest test of global constraint
consistency. Each problem is paired with a verifiable ground-truth
expression. This ground truth supports the QA checks of Section~\ref{sec:qa}
and the binary verification reward of Section~\ref{sec:exp_setup}. The retained
papers yield $4{,}027$ problems after the QA filter. Format-level prompt
templates are in Appendix~\ref{app:prompt_construction}.

\subsection{Quality Assurance and Verifiability}
\label{sec:qa}

Quality assurance combines a two-tier review pipeline (GPT-4o rubric
$\to$ expert validation) with three construction-time choices that bound
the residual error of the automated grader.

\noindent\textbf{Tier 1: Automated Evaluation.}
Each generated problem is scored by GPT-4o along six dimensions:
question clarity, background relevance, answer accuracy, technical
appropriateness, mathematical rigour, and wireless relevance. Scores are assigned on a $5$-point rubric ($1$
invalid, $2$ poor, $3$ acceptable, $4$ good, $5$ excellent). Problems
scoring below $3$ are routed back for revision or excluded outright; the
rest proceed to expert review. Full prompts and the rubric definition
are in Appendix~\ref{app:quality_rubric_human}.

\noindent\textbf{Tier 2: Expert Validation.}
Five PhD-student or postdoctoral domain specialists reviewed the
surviving candidates for mathematical rigour, notational consistency,
problem clarity, and wireless-communications relevance. The retained
screening records cover $5{,}205$ candidate question IDs, including all
$4{,}027$ released problems, for an acceptance rate of
$4{,}027/5{,}205=77.4\%$. The pass was a screening-and-revision process, not
a fully crossed annotation study, so we use its records as provenance rather
than as a source of high-reliability labels: agreement on binary inclusion
is modest (Cohen's $\kappa=0.164$), while $79.6\%$ of rater pairs fall within
one rubric point and $45.2\%$ of include/exclude disagreements are the
adjacent 2-versus-3 pair at the inclusion cut
(Appendix~\ref{app:expert_agreement}).

\noindent\textbf{Verifiability as a Construction-Time Property.}
Every problem carries a reference answer, and a semantic judge
(GPT-4.1-mini) grades each response against it: the judge accepts notation
differences and re-labelling when the expressions are mathematically
equivalent, and rejects sign errors, dropped terms, and wrong dimensions. A
rule-based exact-format scorer runs as a separate lower-bound check. On a
two-stage grader, an early cross-judge probe
(Appendix~\ref{app:evaluation_robustness_app}) finds a judge sensitivity of
$3.4$--$9.3\%$ absolute in the per-model worst case. Comparisons in Section~\ref{sec:exp} hold under one
judge across all rows.

\noindent\textbf{Construction-model dual roles.}
Two models used in construction also appear as evaluated rows:
DeepSeek-R1 (extraction) and GPT-4o (filtering and rubric scoring).
Extraction produces the problem text every model sees; the ground-truth
equation comes from the source paper. The residual channel is selection
and style bias toward content that R1 can extract; the retained provenance
does not quantify the size of this bias. We therefore mark R1 and GPT-4o as
construction-involved calibration rows and exclude them from comparative
claims; the datasheet flags both, together with the dual role of the
GPT-4.1-mini judge.

\subsection{Statistics, Coverage, and Split}
\label{sec:stats}

\begin{figure}[t]
\centering
\includegraphics[width=\linewidth]{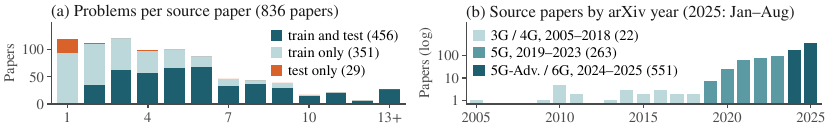}\\[3pt]
\definecolor{fmtMCQ}{HTML}{3A6EA5}\definecolor{fmtFill}{HTML}{1D6F78}\definecolor{fmtFEC}{HTML}{8E1F56}
\tcbset{exbox/.style={enhanced,colback=white,colframe=#1,colbacktitle=#1,coltitle=white,colbacklower=#1!6,
  boxrule=0.4pt,arc=1.5pt,left=2.5pt,right=2.5pt,top=1.5pt,bottom=1.5pt,toptitle=0.6pt,bottomtitle=0.6pt,
  middle=1.2pt,segmentation style={#1!40,line width=0.4pt},
  fonttitle=\scriptsize\bfseries,fontupper=\scriptsize,fontlower=\scriptsize,
  before upper={\raggedright},space to upper,valign upper=center,equal height group=wmbex}}
\makebox[\linewidth][l]{\fontsize{8}{9.6}\selectfont (c) One released test item per format}\\[2pt]
\begin{minipage}[t]{.328\linewidth}
\begin{tcolorbox}[exbox=fmtMCQ,title={MCQ\hfill\mdseries arXiv:2507.12235}]
\emph{Which expression gives the correct theoretical RCS of a PEC
sphere\ldots?}\par\vspace{1.5pt}
\centerline{$\sigma_{sph} = [\textsc{mask}]$}\par\vspace{1.5pt}
\centerline{\begin{tabular}{@{}l@{\hspace{1.2em}}l@{}}(A)~$4\pi R^2$ & (B)~$\pi R^2$\\ (C)~$4\pi R^2/3$ & (D)~$2\pi R^2$\end{tabular}}
\tcblower
\textbf{Answer: (B) $\boldsymbol{\pi R^2}$}\vphantom{$\boldsymbol{\sigma_{sig}^2}$}
\end{tcolorbox}
\end{minipage}\hfill
\begin{minipage}[t]{.328\linewidth}
\begin{tcolorbox}[exbox=fmtFill,title={Fill-in, 50\,\%\hfill\mdseries arXiv:2411.10259}]
\emph{What is the denominator of the expression inside the parentheses?}\par\vspace{1.5pt}
\centerline{$\tau = \left(\frac{\lambda}{[\textsc{mask}]}\right)^{2}$}
\tcblower
\textbf{Answer: $\boldsymbol{4\pi d}$}\vphantom{$\boldsymbol{\sigma_{sig}^2}$}
\end{tcolorbox}
\end{minipage}\hfill
\begin{minipage}[t]{.328\linewidth}
\begin{tcolorbox}[exbox=fmtFEC,title={FEC\hfill\mdseries arXiv:2111.15301}]
\emph{Write the complete equation for the total noise variance.}\par\vspace{1.5pt}
\centerline{$[\textsc{mask}] = [\textsc{mask}]$}
\tcblower
\textbf{Answer: $\boldsymbol{\sigma_t^2 = \sigma_{pr}^2 + \sigma_{bn}^2 + \sigma_{sig}^2}$}
\end{tcolorbox}
\end{minipage}
\caption{\textbf{WirelessMathBench-XL at a glance.} (a) Problems per source
paper, coloured by the splits each paper feeds; (b) source papers by arXiv
year and technology era; (c) one released CC~BY~4.0 test item per format
(background text omitted).}
\label{fig:benchmark_overview}
\end{figure}

WirelessMathBench-XL contains $4{,}027$ problems from $836$ retained
source papers across $20$ technical subfields, selected from a
$970$-paper source pool. For comparison, WirelessMathBench~\citep{li2025wirelessmathbench}
contains $587$ problems from $40$ papers.

\noindent\textbf{Technical coverage.}
Subfield representation reflects the activity profile of the source
literature rather than uniform sampling. Tagged over the $970$-paper source
pool, papers carry $1{,}845$ subfield tags in total (several per paper).
Deep-learning approaches to wireless problems lead ($259$ tags, $14.0\%$),
followed by convex optimisation ($206$, $11.2\%$), MIMO / Massive MIMO
($192$, $10.4\%$), beamforming ($185$, $10.0\%$), and RIS / IRS ($156$,
$8.5\%$). Edge computing ($118$), reinforcement learning ($101$),
information theory ($75$), ISAC ($73$), and OFDM / OFDMA ($70$) follow;
the remaining $10$ subfields, from UAV communications ($69$) to cell-free
networks ($21$), form the long tail, which supplies stress cases for rarer
constraint classes such as the unstated mutual-information constraints of
semantic communications ($32$).

\noindent\textbf{Temporal distribution.}
The retained papers (Figure~\ref{fig:benchmark_overview}b) span three
technological generations: 3G / 4G era (2005--2018: $22$ retained papers, $2.6\%$), 5G deployment
(2019--2023: $263$ retained papers, $31.5\%$), and 5G-Advanced / 6G
research (2024--2025: $551$ retained papers, $65.9\%$). The skew toward
recent work exposes the constraint vocabulary of contemporary wireless
research, while the older slice keeps canonical Shannon-capacity-style
derivations in scope.

\noindent\textbf{Train / test split and source-paper clustering.}
The release ships a problem-level $3{,}227/800$ split rather than a
source-paper-disjoint split: $456$ of the $836$ retained source papers span
both partitions (Figure~\ref{fig:benchmark_overview}a). Of the $800$ test
items, $766$ share a source paper with training and $34$ from $29$ papers do
not, so training rows are release-split baselines. Every record includes
\texttt{paper\_id}. The release materialises the $34$-item \texttt{paper\_disjoint\_test} view
(identifiers for all $34$, text for the $13$ whose source licenses permit
redistribution) and a seed-$0$ grouped assignment with $3{,}222/805$ records and zero source-paper overlap for users
who need to rerun training and evaluation under clustered partitions. The
sensitivity slice is reported in Section~\ref{sec:exp_grpo} and
Appendix~\ref{app:paper-disjoint}. Table~\ref{tab:format_split} gives the
per-format split, and Figure~\ref{fig:benchmark_overview}c shows one released
item per format.

\begin{table}[t]
\centering
\caption{Problems per task format in the release split.}
\label{tab:format_split}
\small
\begin{tabular*}{\linewidth}{@{\extracolsep{\fill}}lrrr@{}}
\toprule
\textbf{Format} & \textbf{Train} & \textbf{Test} & \textbf{Total} \\
\midrule
MCQ & 551 & 133 & 684 \\
Fill-in ($25/50/75\%$ masked) & 1{,}925 ($345/680/900$) & 476 & 2{,}401 \\
FEC & 751 & 191 & 942 \\
\midrule
Total & 3{,}227 & 800 & 4{,}027 \\
\bottomrule
\end{tabular*}
\end{table}

\subsection{Reverse-Probe Contamination Audit}
\label{sec:contamination-audit}

For each problem, we extract $13$-grams from the released prompt surface
(background, question text, and masked equation) after lightweight
LaTeX-token normalisation, index these problem-side grams, stream the
reference corpus, and accumulate per-problem hit counts.

\noindent\textbf{Benchmark-side indexing.} Indexing the
benchmark side makes memory scale with the number of problems rather than
with the number of corpus grams. With $16$ workers and about $6$\,GB of
resident memory, one workstation streams $12.9$\,B RedPajama-arXiv
$13$-grams against all $4{,}027$ problems, and a holder of a private corpus
can run the same code without disclosing it. Hashing,
parallelism, and schema details are in Appendix~\ref{app:audit-method}. The
audit input excludes the unmasked \texttt{correct\_answer} and the
source-paper derivation paragraphs, so $\mathcal{S}_0$ is a label for lexical
overlap on the prompt surface rather than a certificate about exposure to the
target equation.
The audit is independent of the problem-level split: a paper-disjoint split
controls overlap inside the release, not pretraining exposure.

\noindent\textbf{Corpora.} The release-defining target is
RedPajama-arXiv~\citep{weber2024redpajama}: $1{,}558{,}306$ records in our
$100$-shard snapshot, latest timestamp 7 March 2023. Dolma~v1.7
\texttt{arxiv} and ProofPile-2 \texttt{arxiv} turn out to repackage the same
RedPajama dump: Dolma serves it under URLs that match the RedPajama manifest
one-to-one, and ProofPile-2 ships the same documents in the same order with
the same \texttt{arxiv\_id}s, so a dual-slice audit returns identical hits for
every problem (Appendix~\ref{app:audit-method}). These widely used arXiv
pretraining slices therefore provide one source of in-domain evidence, not three.
OpenWebMath serves as an off-domain corroboration run: it flags $16$ of the
$4{,}027$ problems ($0.40\%$), and $99.87\%$ of RedPajama-$\mathcal{S}_0$
problems also have zero OpenWebMath hits. An
off-domain algebraic-stack run flags $2$ of the $4{,}027$ problems ($0.05\%$)
as a pipeline check.

\noindent\textbf{Threshold.} We fix $n=13$ as an operational threshold,
not a validated optimum, following exact-overlap decontamination practice
\citep{brown2020language}, and calibrate it retrospectively. Source papers
dated April 2023 or later cannot match their own source text, so their flag
rate is a collision-rate proxy (not a general false-positive estimate:
related papers and shared phrasing can still match). On the test split the
proxy falls from $64.5\%$ at $n=8$ to $3.1\%$ at $n=13$, the smallest tested
$n$ with a proxy rate at or below $3.1\%$ (Table~\ref{tab:ngram_sweep}). The threshold was fixed
before any full-versus-$\mathcal{S}_0$ result was examined, and the release
exposes memberships at every tested $n$ for all problems so users can replace it.

\noindent\textbf{Cleaned subsets.} Subset definitions and fallback criteria
were fixed before the audit ran; the fallback was not triggered. The strict
subset $\mathcal{S}_0$ holds problems with zero detected RedPajama-arXiv
matches, $\lvert\mathcal{S}_0\rvert=3{,}853$ ($95.7\%$); the lenient subset
$\mathcal{S}_1$ allows up to two distinct source documents,
$\lvert\mathcal{S}_1\rvert=3{,}985$ ($99.0\%$). $\mathcal{S}_0$ is a
fixed-threshold label: on the test split its coverage rises from $33.0\%$ at
$n=8$ to $98.8\%$ at $n=15$.

\begin{table}[t]
\centering
\caption{Strict zero-hit label versus $n$-gram length on the $800$-item test
split. Post-/Pre-snap.\ are flag rates for papers dated after/before the
RedPajama-arXiv snapshot; post-snapshot papers cannot match their own source.}
\label{tab:ngram_sweep}
\small
\begin{tabular*}{\linewidth}{@{\extracolsep{\fill}}rrrrrr@{}}
\toprule
$n$ & $\mathcal{S}_0$ count & $\mathcal{S}_0$ \% & Flip vs.\ $13$ (\%) & Post-snap.\ \% & Pre-snap.\ \% \\
\midrule
$8$  & $264/800$ & $33.0$ & $63.2$ & $64.5$ & $74.9$ \\
$10$ & $600/800$ & $75.0$ & $21.2$ & $23.6$ & $29.2$ \\
$12$ & $745/800$ & $93.1$ & $3.1$  & $6.8$  & $7.2$ \\
$13$ & $770/800$ & $96.2$ & $0.0$  & $3.1$  & $5.6$ \\
$15$ & $790/800$ & $98.8$ & $2.5$  & $1.3$  & $1.0$ \\
\bottomrule
\end{tabular*}
\end{table}

\subsection{Implicit-Constraint Taxonomy}
\label{sec:taxonomy}

The load-bearing content of a wireless-math problem is often the implicit
constraint, not only the equation. Four recurring classes informed
distractor construction and masking design:
\textbf{(C1) power non-negativity} (KKT closed-forms admitting negative
roots that need a water-filling $(\cdot)^{+}$ never stated in the text);
\textbf{(C2) complex-matrix dimensionality} (silent
$\mathbf{H}^{H}\mathbf{H}\!\leftrightarrow\!\mathbf{H}\mathbf{H}^{H}$
swaps under Hermitian transpose, Kronecker product, inversion);
\textbf{(C3) log-domain validity} ($\log(1+\gamma)$ used outside its
domain after aggressive interference cancellation); and \textbf{(C4)
signal causality} (filters or precoders conditioning on
$\mathbf{h}[t+1]$ inside a $t$-time decision).\label{sec:taxonomy_classes}
The classes are descriptive in this release (no per-instance labels); full
definitions, the MCQ$\to$Fill-in$\to$FEC probing structure, and the sampling
protocol for future prevalence labels are in Appendix~\ref{app:taxonomy}.

\section{Experiments}
\label{sec:exp}

We evaluate three properties of the release: how hard the benchmark is and
whether its training split is learnable (Section~\ref{sec:exp_main}), how
scores respond to the audit (Section~\ref{sec:exp_contam}), and whether
wireless training regresses general skills (Section~\ref{sec:exp_transfer}).

\subsection{Evaluation Setup}
\label{sec:exp_setup}

All rows use the $800$-problem release test split, and the GPT-4.1-mini
judge of Section~\ref{sec:qa} grades every response. Generation uses $T=0.6$ and
top-$p=1.0$. Open-weight $7$B-scale and smaller models run locally under
\texttt{vllm 0.8.5}, and hosted models share one OpenRouter request stack
(Appendix~\ref{app:cleaned_detail}). Base-model GRPO checkpoints (the WirelessMathLM family and
DeepSeekMath-$7$B-RL) are queried through the raw-prompt
\texttt{/v1/completions} endpoint to match training-time prompting; chat
models use the chat endpoint. Because token budget materially affects
frontier reasoning models, we report five frontier rows at a $16$k answer
budget and fourteen rows at a locked $2$k budget (the $0.5$B row is in
Appendix~\ref{app:cleaned_detail}); full and $\mathcal{S}_0$
scores within a row share the budget, so audit comparisons are paired.
The split follows from the budget sensitivity of reasoning models:
Gemini-$2.5$-Pro scores $41.25\%$ at $2$k and $86.88\%$ at $16$k, because at
$2$k its visible responses average $435$ characters and $21/800$ are empty;
GPT-$5$-mini gains $3.00$\,pp from the same change.
The training reward shares the GPT-4.1-mini verifier family, so WirelessMathLM
rows measure learnability under the published verifier stack.

\noindent\textbf{GRPO recipe.} WirelessMathLM is trained with Group Relative
Policy Optimisation~\citep{shao2024deepseekmath} from base checkpoints
(no SFT warm-start) under the reward
$r(x,y) = 0.1\,r_{\text{format}}(y) + 0.9\,r_{\text{accuracy}}(x,y)$
for $40$ epochs ($240$ steps), AdamW at $10^{-6}$, cosine annealing, and KL
penalty $\beta = 0.01$, on $4\times$ NVIDIA RTX 4090 D ($28.4$\,h, $53.4$\,h
for $3$\,B, $7$\,B); hyperparameters are in
Appendix~\ref{app:training_details}.

\subsection{Main Results: Difficulty and Learnability}
\label{sec:exp_main}
\label{sec:exp_grpo}

\begin{table}[t]
\centering
\caption{Accuracy (\%) on the $800$-item test split, the $770$-item strict
zero-hit subset $\mathcal{S}_0$, and the $310$-item public subset, with
$\Delta=\mathrm{Full}-\mathcal{S}_0$ and paired-bootstrap $95\%$ CIs.
$^{\dagger}$Construction-involved rows, reported descriptively.}
\label{tab:frontier_calibration}
\label{tab:main_results}
\small
\setlength{\tabcolsep}{0pt}
\begin{tabular*}{\linewidth}{@{\extracolsep{\fill}}lcrrrr@{}}
\toprule
\textbf{Model} & \textbf{Size} & \textbf{Full} & \textbf{$\mathcal{S}_0$} & \textbf{$\Delta$\,(pp) [$95\%$ CI]} & \textbf{Public} \\
\midrule
\multicolumn{6}{@{}l}{\textit{Frontier models, $16$k answer budget}} \\
GPT-$5$-mini                  & --   & 91.25 & 91.69 & $-0.44$ $[-1.05, +0.07]$ & 90.32 \\
GPT-$5.2$-chat                & --   & 90.75 & 90.91 & $-0.16$ $[-0.66, +0.25]$ & 90.00 \\
GPT-$4.1$-mini                & --   & 89.62 & 89.35 & $+0.27$ $[-0.04, +0.51]$ & 89.35 \\
Gemini-$2.5$-Pro              & --   & 86.88 & 87.01 & $-0.14$ $[-0.69, +0.32]$ & 85.48 \\
Gemini-$2.5$-Flash            & --   & 86.50 & 86.49 & $+0.01$ $[-0.51, +0.43]$ & 84.19 \\
\midrule
\multicolumn{6}{@{}l}{\textit{General-purpose models, locked $2$k budget}} \\
Grok-$4$-Fast                 & --   & 88.00 & 88.31 & $-0.31$ $[-0.90, +0.19]$ & 84.84 \\
DeepSeek-V$3.1$               & 671B & 84.88 & 84.42 & $+0.46$ $[+0.14, +0.75]$ & 83.87 \\
Claude-$4.0$-Sonnet           & --   & 83.00 & 82.73 & $+0.27$ $[-0.20, +0.67]$ & 82.26 \\
o$4$-mini                     & --   & 81.88 & 82.08 & $-0.20$ $[-0.83, +0.35]$ & 79.35 \\
DeepSeek-R$1^{\dagger}$       & 671B & 79.12 & 79.09 & $+0.03$ $[-0.55, +0.55]$ & 77.74 \\
GPT-$4$o$^{\dagger}$          & --   & 77.00 & 76.88 & $+0.12$ $[-0.48, +0.63]$ & 75.81 \\
Qwen$2.5$-$72$B-Instruct      & 72B  & 73.62 & 73.77 & $-0.14$ $[-0.80, +0.46]$ & 73.87 \\
Llama-$3.3$-$70$B-Instruct    & 70B  & 61.50 & 61.56 & $-0.06$ $[-0.74, +0.60]$ & 59.35 \\
GPT-$5$-nano                  & --   & 52.50 & 52.86 & $-0.36$ $[-1.07, +0.33]$ & 51.29 \\
\midrule
\multicolumn{6}{@{}l}{\textit{Open $7$B-scale models, locked $2$k budget}} \\
WirelessMathLM-$7$B (ours)    & 7B   & 47.88 & 48.70 & $-0.83$ $[-1.51, -0.17]$ & 46.45 \\
DeepSeekMath-$7$B-RL          & 7B   & 41.00 & 40.91 & $+0.09$ $[-0.58, +0.78]$ & 38.71 \\
Qwen$2.5$-Math-$7$B-Instruct  & 7B   & 40.00 & 40.13 & $-0.13$ $[-0.77, +0.55]$ & 40.65 \\
WirelessMathLM-$3$B (ours)    & 3B   & 25.50 & 26.36 & $-0.86$ $[-1.28, -0.46]$ & 25.16 \\
\bottomrule
\end{tabular*}
\end{table}

\begin{figure}[t]
\centering
\includegraphics[width=\linewidth]{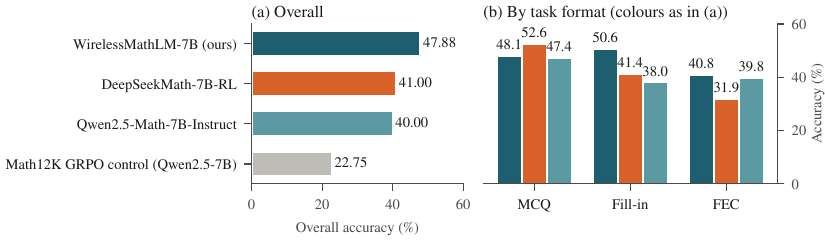}
\caption{Open $7$B-scale models on the test split (locked $2$k). (a) Overall
accuracy, including a Qwen$2.5$-$7$B GRPO control trained on Math12K;
(b) accuracy by task format.}
\label{fig:learnability}
\end{figure}

Table~\ref{tab:frontier_calibration} separates the field. Frontier models
reach $86.5$--$91.3\%$ at a $16$k budget, general-purpose models span
$52.5$--$88.0\%$ at $2$k, and open $7$B-scale models stay at $40$--$48\%$.
The frontier rows form one high-accuracy cluster rather than a resolved
ranking: the matched-budget GPT-$5.2$-chat versus GPT-$5$-mini difference is
$-0.50$\,pp $[-2.88,+1.88]$. Construction-involved rows (DeepSeek-R1,
GPT-4o) are kept as descriptive calibration and excluded from comparative
claims. The Public column re-scores every row on the $310$-item public test
subset from the same archived verdicts and is the externally reproducible
view, since $490$ test items are withheld under their source-paper licenses
(Appendix~\ref{app:public_subset}). The ranking carries over to it: no
pairwise ordering among the five frontier rows changes, and one of the $91$
pairs among the $14$ locked-$2$k rows swaps, namely DeepSeekMath-$7$B-RL and
Qwen$2.5$-Math-$7$B-Instruct, which are $1.00$\,pp apart on the full split.

\noindent\textbf{Formats separate what overall accuracy merges.}
Table~\ref{tab:per_format_main} splits the locked-$2$k rows by task format.
DeepSeekMath-$7$B-RL and Qwen$2.5$-Math-$7$B-Instruct are $1.00$\,pp apart
overall, yet the first leads by $5.26$\,pp on MCQ and the second by
$7.85$\,pp on FEC; GPT-$5$-nano reaches $66.92\%$ on MCQ but $36.65\%$ on
FEC, where only definitions, assumptions, and boundary conditions are given.
This is the separation the formats were designed for: deliberately different
difficulty profiles (Section~\ref{sec:extraction}).

\begin{table}[t]
\centering
\caption{Accuracy (\%) by task format for the locked-$2$k rows of
Table~\ref{tab:main_results}; Fill-in pools the $25/50/75\%$ rungs.}
\label{tab:per_format_main}
\small
\begin{tabular*}{\linewidth}{@{\extracolsep{\fill}}lrrrr@{}}
\toprule
\textbf{Model} & \textbf{MCQ} & \textbf{Fill-in} & \textbf{FEC} & \textbf{Overall} \\
\midrule
Grok-$4$-Fast                 & 87.22 & 88.24 & 87.96 & 88.00 \\
DeepSeek-V$3.1$               & 85.71 & 85.71 & 82.20 & 84.88 \\
Claude-$4.0$-Sonnet           & 84.21 & 84.03 & 79.58 & 83.00 \\
o$4$-mini                     & 83.46 & 83.82 & 75.92 & 81.88 \\
DeepSeek-R$1^{\dagger}$       & 81.20 & 80.88 & 73.30 & 79.12 \\
GPT-$4$o$^{\dagger}$          & 79.70 & 77.73 & 73.30 & 77.00 \\
Qwen$2.5$-$72$B-Instruct      & 70.68 & 75.42 & 71.20 & 73.62 \\
Llama-$3.3$-$70$B-Instruct    & 71.43 & 57.77 & 63.87 & 61.50 \\
GPT-$5$-nano                  & 66.92 & 54.83 & 36.65 & 52.50 \\
\midrule
WirelessMathLM-$7$B (ours)    & 48.12 & 50.63 & 40.84 & 47.88 \\
DeepSeekMath-$7$B-RL          & 52.63 & 41.39 & 31.94 & 41.00 \\
Qwen$2.5$-Math-$7$B-Instruct  & 47.37 & 38.03 & 39.79 & 40.00 \\
WirelessMathLM-$3$B (ours)    & 29.32 & 28.15 & 16.23 & 25.50 \\
\bottomrule
\end{tabular*}
\end{table}

\noindent\textbf{The training split is learnable.} WirelessMathLM-$7$B
reaches $47.88\%$, above the two disclosed $7$B math specialists,
DeepSeekMath-$7$B-RL ($41.00\%$; paired $95\%$ CI under the same judge
$[+2.88,+11.00]$) and Qwen$2.5$-Math-$7$B-Instruct ($40.00\%$;
$[+3.88,+11.88]$); it leads on Fill-in, is ahead of DeepSeekMath-$7$B-RL and level
with Qwen$2.5$-Math-$7$B-Instruct on FEC, and trails DeepSeekMath-$7$B-RL on
MCQ (Figure~\ref{fig:learnability}). On $\mathcal{S}_0$ it comes within
$4.16$\,pp of GPT-$5$-nano ($95\%$ CI $[-0.13,+8.31]$;
Appendix~\ref{app:cleaned_detail}). Under the training-time evaluation,
GRPO raises its point estimate on $19$ of the $20$ subdomain tags of the test
split (Appendix~\ref{app:per_subdomain}).

\noindent\textbf{Size-matched control.} We compare training data at
matched size and steps. Trained with GRPO on $3{,}227$ Math12K examples
for $40$ epochs ($240$ steps), the same Qwen$2.5$-$7$B base reaches
$22.75\%$, so the released split adds $25.13$\,pp ($95\%$ item-bootstrap CI
$[21.13, 29.00]$; one run per condition; batch, prompt-template, and reward
implementations are not identical). This shows domain-aligned learnability
on the release split; it does not establish that wireless mathematics is
intrinsically harder than other mathematics or causally isolate wireless
constraints. Appendix~\ref{app:training_details} reports training-time checks across
Qwen$2.5$-$0.5$B/$3$B/$7$B and Qwen$3$-$4$B, together with an SFT ablation.

\noindent\textbf{Unseen source papers.} $766$ of the $800$ test items share
a source paper with training. On the $34$ items whose source papers are
absent from training, WirelessMathLM-$7$B scores $58.8\%$ versus $47.4\%$ on
the rest (Table~\ref{tab:paper_disjoint_sensitivity}), so we see no drop on
unseen papers; the slice is small and its intervals are wide.

\begin{table}[t]
\centering
\caption{Accuracy (\%) on the $34$ test items from source papers absent from
training (PD) and on the remaining $766$ (Rest), locked $2$k
(Appendix~\ref{app:cleaned_detail}).}
\label{tab:paper_disjoint_sensitivity}
\small
\begin{tabular*}{\linewidth}{@{\extracolsep{\fill}}lrrr@{}}
\toprule
\textbf{Model} & \textbf{PD} & \textbf{Rest} & \textbf{$\Delta$\,(pp) [$95\%$ CI]} \\
\midrule
GPT-$5$-mini ($2$k)          & 97.1 & 87.9 & $+9.2$ $[+2.3,+14.0]$ \\
Grok-$4$-Fast                & 82.4 & 88.3 & $-5.9$ $[-19.8,+6.4]$ \\
DeepSeek-V$3.1$              & 85.3 & 84.9 & $+0.4$ $[-12.5,+11.6]$ \\
WirelessMathLM-$7$B (ours)   & 58.8 & 47.4 & $+11.4$ $[-5.6,+28.4]$ \\
DeepSeekMath-$7$B-RL         & 58.8 & 40.2 & $+18.6$ $[+1.5,+35.2]$ \\
Qwen$2.5$-Math-$7$B-Instruct & 38.2 & 40.1 & $-1.8$ $[-18.3,+15.5]$ \\
\bottomrule
\end{tabular*}
\end{table}

\subsection{Audit-Conditioned Evaluation}
\label{sec:exp_contam}

Only $30$ of the $800$ test items carry detected overlap. If all
flagged items were answered correctly, the largest possible positive $\Delta$
would range from $0.31$ to $0.51$\,pp across the five frontier models.
Full-versus-$\mathcal{S}_0$ is therefore a bounded
stability summary for one channel, exact lexical overlap between the prompt
surface and RedPajama-arXiv at $n=13$, and is structurally underpowered as a
test of contamination effects. Exposure through target answers, paraphrases,
alternate versions, closed corpora, post-training, or reward-verifier
co-adaptation lies outside this channel.

Within the channel, $\Delta$ lies in $[-0.86,+0.46]$\,pp across all rows
($10{,}000$ paired bootstrap resamples by problem id), and every frontier row
moves by at most $0.44$\,pp. Within the channel, no pairwise ordering of point
estimates changes sign, neither among the frontier rows nor among the $153$ pairs in the locked
$2$k table of $18$ models (Appendix~\ref{app:cleaned_detail}). A
stricter exact-format scorer (MCQ letter match plus exact comparison of
extracted \texttt{\textbackslash boxed} answers) gives deltas in
$[-0.21,+0.49]$\,pp for the frontier and $7$B-scale rows. The five overlap
buckets ($770/11/9/4/6$ items) show $|\rho|\leq0.11$ in every bucket-level
Spearman check, and every locked-row $\Delta(\text{Full}-\mathcal{S}_1)$ lies
in $[-0.36,+0.15]$\,pp. All intervals that lie entirely below zero belong
to WirelessMathLM checkpoints ($-0.83$, $-0.86$, and $-0.08$\,pp at $7$B,
$3$B, and $0.5$B), which score slightly higher on $\mathcal{S}_0$; DeepSeek-V$3.1$
($+0.46$\,pp) has the only interval entirely above zero.

\begin{table}[t]
\centering
\caption{$\Delta=\mathrm{Full}-\mathcal{S}_0^{(n)}$ (pp) for the $16$k frontier
rows when the zero-hit label is recomputed at each $n$-gram length.}
\label{tab:frontier_ngram_delta}
\small
\begin{tabular*}{\linewidth}{@{\extracolsep{\fill}}rrrrrrr@{}}
\toprule
$n$ & $|\mathcal{S}_0^{(n)}|$ & GPT-$5$-mini & GPT-$5.2$-chat & GPT-$4.1$-mini & Gem.-$2.5$-Pro & Gem.-$2.5$-Flash \\
\midrule
8  & 264 & $-0.04$ & $+2.49$ & $+1.75$ & $+1.27$ & $+1.27$ \\
10 & 600 & $-0.42$ & $-0.75$ & $+0.96$ & $-0.62$ & $+0.17$ \\
12 & 745 & $-0.29$ & $-0.26$ & $+0.50$ & $-0.10$ & $+0.19$ \\
13 & 770 & $-0.44$ & $-0.16$ & $+0.27$ & $-0.14$ & $+0.01$ \\
15 & 790 & $-0.14$ & $-0.14$ & $+0.00$ & $+0.04$ & $+0.04$ \\
\bottomrule
\end{tabular*}
\end{table}

The threshold matters more than the model set. When we recompute the frontier
comparisons on the same verdicts at $n\in\{8,10,12,15\}$
(Table~\ref{tab:frontier_ngram_delta}), the aggressive $n=8$ label keeps
$264/800$ items and raises the largest delta to $+2.49$\,pp, while for
$n\geq10$ every frontier delta stays within $0.96$\,pp. Across the full
split and all five thresholds, none of the ten frontier pairwise orderings
changes sign. A narrow answer-surface check at $n=13$ finds $3/800$
RedPajama-arXiv document hits and moves the frontier rows by at most
$0.05$\,pp.

\noindent\textbf{Portability.} The same code runs unchanged on other
benchmarks. On the $587$-item WirelessMathBench predecessor it returns zero
detected hits (its prompts expose about $29$ probe $13$-grams per item, versus
about $127$ for WirelessMathBench-XL), and on MATH-500 it flags $5/500$ items.
Each flag ships with its matched strings and source-document counts, so the
largest one ($78$ matches across $12$ documents) can be traced to a generic
arithmetic sequence (Appendix~\ref{app:audit:external-reruns}).

\subsection{Regression Checks Beyond Wireless}
\label{sec:exp_transfer}

Wireless GRPO training does not cause obvious regressions outside the
domain. On four general-purpose benchmarks (Table~\ref{tab:general_benchmarks}),
every WirelessMathLM delta is positive except IFEval at $7$B ($-0.39$), with
the largest gains on MMLU ($+11.90$) and GPQA ($+10.10$) at $7$B. On five
general-math benchmarks (Table~\ref{tab:transfer_learning}), which contribute
no training problems, the $7$B and $3$B checkpoints gain $+5.40$ and $+6.52$
average accuracy, led by MATH-$500$ ($+8.93$ and $+16.60$). The $7$B GRPO row averages
three evaluation runs. These rows are regression checks.

\begin{table}[t]
\centering
\caption{Accuracy (\%) on general-purpose benchmarks before and after
WirelessMathLM GRPO.}
\label{tab:general_benchmarks}
\small
\begin{tabular*}{\linewidth}{@{\extracolsep{\fill}}lrrrr@{}}
\toprule
\textbf{Model} & \textbf{MMLU} & \textbf{GPQA} & \textbf{IFEval} & \textbf{HumanEval} \\
\midrule
Qwen$2.5$-$3$B-Base & 42.49 & 23.74 & 26.99 & 61.59 \\
\quad + GRPO (WirelessMathLM-$3$B) & 45.30 & 28.28 & 31.42 & 65.24 \\
\quad $\Delta$      & $+2.81$ & $+4.54$ & $+4.43$ & $+3.65$ \\
\midrule
Qwen$2.5$-$7$B-Base & 38.84 & 28.79 & 36.06 & 63.41 \\
\quad + GRPO (WirelessMathLM-$7$B) & 50.74 & 38.89 & 35.67 & 64.63 \\
\quad $\Delta$      & $+11.90$ & $+10.10$ & $-0.39$ & $+1.22$ \\
\bottomrule
\end{tabular*}
\end{table}

\begin{table}[t]
\centering
\caption{Accuracy (\%) on general-math benchmarks before and after
WirelessMathLM GRPO.}
\label{tab:transfer_learning}
\small
\begin{tabular*}{\linewidth}{@{\extracolsep{\fill}}lrrrrrr@{}}
\toprule
\textbf{Model} & \textbf{MATH-$500$} & \textbf{Minerva} & \textbf{Olympiad} & \textbf{AMC} & \textbf{AIME-$24$} & \textbf{Avg.} \\
\midrule
Qwen$2.5$-$7$B-Base & 52.00 & 12.13 & 21.93 & 27.71 &  6.67 & 24.09 \\
\quad + GRPO        & 60.93 & 12.62 & 27.36 & 36.55 & 10.00 & 29.49 \\
\quad $\Delta$      & $+8.93$ & $+0.49$ & $+5.43$ & $+8.84$ & $+3.33$ & $+5.40$ \\
\midrule
Qwen$2.5$-$3$B-Base & 41.60 &  5.88 & 14.67 & 18.07 &  0.00 & 16.04 \\
\quad + GRPO        & 58.20 &  9.93 & 22.96 & 21.69 &  0.00 & 22.56 \\
\quad $\Delta$      & $+16.60$ & $+4.05$ & $+8.29$ & $+3.62$ & $0.00$ & $+6.52$ \\
\bottomrule
\end{tabular*}
\end{table}

\section{Related Work}
\label{sec:related}

\noindent\textbf{Contamination and benchmark validity.}
Memorisation scales with model size and training-set duplication
\citep{carlini2023quantifying}, motivating per-benchmark contamination
measurement \citep{sainz2023nlp,xu2024contamsurvey}. Model-side probes
infer exposure without corpus access, including probability-based
pretraining-data detectors, time-split black-box tests, and test-set slot
guessing \citep{shi2023detecting,oren2023proving,golchin2024timetravel,deng2024investigating};
corpus-side systems such as WIMBD provide search/count infrastructure for
large pretraining mixtures \citep{elazar2024wimbd}. Recent harder benchmarks
post-date some model cutoffs or explicitly address benchmark exposure
\citep{sun2025olymmath,mirzadeh2024gsm} but lose that protection after release
and repeated evaluation. Where WIMBD indexes the corpus, our reverse-probe
audit indexes the benchmark, so the evidence ships with the benchmark as
per-problem metadata, and anyone holding a corpus can recompute it on the
distributed records. The audit complements model-side probes, web-scale search, and near-duplicate
retrieval.

\noindent\textbf{Verifiable technical-domain benchmarks.}
General math suites such as GSM8K, MATH, OlympiadBench, NuminaMath,
MathOdyssey, and OlymMATH
\citep{cobbe2021training,hendrycks2021measuring,he2024olympiadbench,li2024numinamath,fang2024mathodyssey,sun2025olymmath}
stress symbolic reasoning but do not target wireless implicit
constraints. Telecom LLM resources
\citep{shao2024wirelessllm,zou2024telecomgpt,maatouk2024tele,maatouk2023teleqna}
mostly evaluate QA or assistant behaviour; the closest wireless-math
resource is the $40$-paper WirelessMathBench~\citep{li2025wirelessmathbench}.
WirelessMathBench-XL extends this line with source-paper provenance, expert
acceptance, a published verifier with a rule-based lower-bound check, and per-problem audit
metadata as part of the benchmark definition.

\noindent\textbf{Verifier RL and post-training.}
We use GRPO/verifier RL off the shelf
\citep{shao2024deepseekmath,guo2025deepseek,yang2024qwen2.5-math,yang2025qwen3,openai2024learning}
to show that the release split is learnable; the contribution is the
audited benchmark and its release metadata, not a new optimiser.

\section{Conclusion}
\label{sec:conclusion}

WirelessMathBench-XL makes a rerunnable contamination audit part of the
benchmark release. Its reverse-probe design indexes the benchmark rather than
the corpus, so overlap evidence for each problem ships with the data, and
anyone holding a corpus can recompute it on the distributed records. Applied
to RedPajama-arXiv, the audit also shows that the Dolma and ProofPile-2 arXiv slices repackage the same dump. The
benchmark provides $4{,}027$ wireless-math problems from $836$ papers with
source-paper identifiers, verifier-facing answers, cleaned and paper-disjoint
views, and Croissant/Datasheet metadata. Its maintainers commit to at least
three years of maintenance with audit re-runs against newly released public
corpora. It separates frontier models
($86.5$--$91.3\%$) from open $7$B models ($40$--$48\%$), and its training split
lifts a $7$B model $25.1$\,pp above a size-matched general-math control. Its
three formats separate models that overall accuracy places together
(Table~\ref{tab:per_format_main}), and the same audit code runs unchanged on
other benchmarks, returning zero hits on the $587$-item predecessor and $5/500$
flags on MATH-500.
Within the audited channel every frontier score moves by at most $0.44$\,pp
and no frontier ordering changes, and the frontier ordering also holds on the
$310$-item public test subset that anyone can rerun.

\paragraph{Scope and limitations.}
(i) \emph{Channel.} $\mathcal{S}_0$ labels exact prompt-surface $13$-gram
overlap with one public slice; target-answer, paraphrase and near-duplicate,
alternate-version, post-training, and verifier-co-adaptation exposure are
outside it, and $n=13$ is an operational threshold rather than a validated
optimum. (ii) \emph{Corpus coverage.} RedPajama-arXiv is not the training
mixture of any evaluated proprietary model, and papers after its March 2023
snapshot cannot match their own source, so a zero-hit label applies to this
corpus only; holders of other corpora can run the same audit privately.
(iii) \emph{Power.} With $30/800$ flagged test items, detected overlap can
shift a frontier score by at most $0.31$--$0.51$\,pp, so full-versus-$\mathcal{S}_0$
bounds this channel rather than testing for contamination.
(iv) \emph{Annotation.} Historical expert screening has low agreement
($\kappa=0.164$) and serves as provenance rather than as a reliability
guarantee; a blind, fully crossed expert audit is future work.
(v) \emph{Construction.} DeepSeek-R1 and GPT-4o helped construct the problems
and are reported descriptively; the stochastic construction path is not
bitwise reproducible, while the shipped artifact and its provenance are.
The release split is by problem, so training rows are release-split
learnability checks under the published verifier stack.
(vi) \emph{Distribution.} The $1{,}542$ records whose source licenses permit
redistribution are released under those licenses (code under MIT); full-split
scores use the authors' complete copy, and the $310$-item public test subset
is the externally reproducible view; identifiers, source licenses, and audit
metadata are public for all $4{,}027$ problems.

\begin{ack}
We thank Google for the Gemini API credits provided through the Gemini Academic Program, which supported the Gemini experiments in this work. The authors declare no competing interests.
\end{ack}

\bibliographystyle{plainnat}
\bibliography{references}

\appendix

\section{Dataset Construction Details}
\label{app:data_collection}

\subsection{Paper Collection and Filtering}

The source-paper harvest starts from $24$ arXiv categories chosen to cover core
wireless communications (networking, signal processing, information theory),
adjacent AI/ML work, networked systems, optimisation, and emerging physical-layer
techniques. The first-stage filter combines category membership with keyword
evidence in titles and abstracts. It is intentionally broad: the goal is high
recall for papers that may contain reusable wireless mathematical structure, not
a pure physical-layer taxonomy.

Each candidate receives a deterministic relevance score from title/abstract
keyword matches and category bonuses: keyword matches are weighted $0.6$ in the
title and $0.3$ in the abstract, and membership in \texttt{eess.SP},
\texttt{cs.NI}, or \texttt{cs.IT} adds a bonus of $0.4$, $0.35$, or $0.3$,
respectively. The score is used only to order candidates
for the second-stage verifier; it is not a quality label. The second stage uses
GPT-4o to retain papers with substantial mathematical content, after which the
structured extraction pipeline operates on arXiv source when available and on
PDF-derived text otherwise. PDF-to-markdown conversion uses
MinerU~\citep{wang2024mineruopensourcesolutionprecise} to preserve equations and
LaTeX-like mathematical notation.

The resulting source pool is then passed through the extraction and QA pipeline
described in Section~\ref{sec:dataset_construction}: DeepSeek-R1 proposes
system-model extractions and candidate problems, Tier-1 rubric scoring removes
obvious failures, and Tier-2 domain-specialist review accepts only records whose
variables, constraints, and ground-truth answers are sufficiently specified for
evaluation. This appendix records the filtering logic needed to interpret the
released corpus; exact prompts are provided in Appendix~\ref{app:prompt_construction}.

\section{Prompt Construction for Dataset Generation and Evaluation}
\label{app:prompt_construction}

We employ specialized prompt templates for dataset construction, quality assessment, and standardized evaluation to ensure consistency and fairness across all stages of our methodology.

\subsection{System Model Extraction Prompt}

The following prompt template guides the extraction of mathematical models from research papers:
\begin{tcolorbox}[colback=gray!5,colframe=black!50,title=System Model Extraction Template,breakable]
\begin{footnotesize}

Task: Act as an expert in wireless communications and mathematical modeling. Extract and summarize the mathematical system modeling from the paper.

\vspace{0.5em}
\textbf{STRUCTURE REQUIREMENTS}

\begin{enumerate}
    \item \textbf{Model Extraction}:
    \begin{enumerate}
        \item[a)] Identify ALL system equations with context
        \item[b)] For each equation:
        \begin{enumerate}
            \item[i)] List ALL variables with units/dimensions
            \item[ii)] Specify underlying assumptions
            \item[iii)] Note domain restrictions
        \end{enumerate}
    \end{enumerate}

    \item \textbf{Summary Organization}:
    \begin{itemize}
        \item \texttt{\textbackslash paragraph\{Background\}} (2-3 sentences contextualizing the model)
        \item \texttt{\textbackslash paragraph\{Key Assumptions\}} (bullet points)
        \item \texttt{\textbackslash paragraph\{Parameter Definitions\}} (table-like structure)
        \item \texttt{\textbackslash paragraph\{Core Equations\}} (numbered with original labels)
    \end{itemize}

    \item \textbf{Equation Formatting}:
    \begin{itemize}
        \item Vectors: $\boldsymbol{v}$
        \item Matrices: $\mathbf{M}$
        \item Operators: $\mathrm{diag}, \mathrm{tr}$
        \item Complex numbers: $j$ for imaginary unit
    \end{itemize}
\end{enumerate}

\vspace{0.5em}
\textbf{CONTENT GUIDELINES}

\begin{itemize}
    \item Variable Explanations:
    \begin{itemize}
        \item For each symbol: $\theta$ (Type: Phase shift; Domain: $[0,2\pi)$; Unit: rad)
        \item Matrix dimensions: $\mathbf{H} \in \mathbb{C}^{N\times M}$
        \item Distinguish similar symbols: $h_{ij}$ vs $h_{i}^{(j)}$
    \end{itemize}

    \item Model Validation:
    \begin{itemize}
        \item Verify dimensional consistency
        \item Check boundary conditions
        \item Confirm parameter unit homogeneity
    \end{itemize}
\end{itemize}

\end{footnotesize}
\end{tcolorbox}

\subsection{Question Generation Prompt}

The following template generates exam-style questions from extracted models:

\begin{tcolorbox}[colback=gray!5,colframe=black!50,title=Question Generation Template,breakable]
\begin{footnotesize}

Task: Generate exam-style questions from research paper summaries.

\vspace{0.5em}
\textbf{STRUCTURE REQUIREMENTS}

\begin{enumerate}
    \item \textbf{Per Equation Processing}:
    \begin{enumerate}
        \item[a)] Identify ALL system model equations
        \item[b)] For EACH equation:
        \begin{enumerate}
            \item[i)] Mask the RHS with [MASK]
            \item[ii)] Generate 1 MCQ with 4 plausible options
            \item[iii)] Create 4 progressive fill-in-the-blank subquestions: 25\%, 50\%, 75\%, and 100\% key symbols masked
        \end{enumerate}
    \end{enumerate}

    \item \textbf{Question Components}:
    \begin{itemize}
        \item For MCQs:
        \begin{itemize}
            \item Background: MUST include detailed variable definitions, e.g.,
            ``where $\boldsymbol{x}$ is the transmitted signal vector, $\mathbf{H} \in \mathbb{C}^{N \times M}$ represents the channel matrix\ldots''
            \item Equation: Masked equation in display math mode
            \item Question: Explicitly ask to replace [MASK]
            \item Options: 4 LaTeX-formatted choices (A)--(D)
            \item Answer: Detailed derivation walkthrough
        \end{itemize}
    \end{itemize}
\end{enumerate}

\vspace{0.5em}
\textbf{ENHANCED BACKGROUND REQUIREMENTS}

\begin{itemize}
    \item Variable Definition Format:
    \begin{itemize}
        \item Start with system context: ``In this [type of system]\ldots''
        \item List EVERY symbol that appears in the equation
        \item Include matrix/vector dimensions
        \item Specify units where applicable: ``(in watts)'', ``(in Hz)''
        \item Explain subscripts and superscripts
    \end{itemize}

    \item Distractor Design:
    \begin{enumerate}
        \item Matrix dimension mismatches
        \item Incorrect operator sequences
        \item Missing $\mathrm{diag}()$ operators
        \item Channel matrix transposition errors
        \item Incorrect matrix multiplication order
    \end{enumerate}

    \item Masking Strategy:
    \begin{itemize}
        \item 25\%: Single critical variable
        \item 50\%: Two interdependent terms
        \item 75\%: Multiple components
        \item 100\%: Full equation recall
    \end{itemize}
\end{itemize}

\end{footnotesize}
\end{tcolorbox}

\subsection{Quality Assessment Framework Prompt}

To ensure consistent quality evaluation across the dataset, we employ a comprehensive assessment framework with few-shot learning enhancement.

\begin{tcolorbox}[colback=gray!5,colframe=black!50,title=Quality Assessment Prompt with Few-Shot Learning,breakable]
\begin{footnotesize}

You are an expert evaluator specializing in wireless communication and mathematics education. Your task is to assess the quality of technical questions designed for advanced undergraduate and graduate students in wireless communications.

\textbf{EVALUATION METHODOLOGY.} Follow this systematic approach:

\textit{Step 1: Initial Question Analysis}
\begin{itemize}
    \item Read the question, background, equation, and answer carefully
    \item Identify the technical domain and complexity level
    \item Check for obvious errors or inconsistencies
\end{itemize}

\textit{Step 2: Multi-Dimensional Quality Assessment.} Evaluate each dimension on a 1--5 scale:
\begin{enumerate}
    \item \textbf{Question Clarity (1--5):} Crystal clear vs confusing/incomprehensible
    \item \textbf{Background Relevance (1--5):} Comprehensive context vs inadequate background
    \item \textbf{Answer Accuracy (1--5):} Completely correct vs incorrect/flawed
    \item \textbf{Technical Appropriateness (1--5):} Perfect difficulty vs inappropriate level
    \item \textbf{Mathematical Rigor (1--5):} Excellent notation vs poor rigor
    \item \textbf{Wireless Relevance (1--5):} Highly relevant vs not relevant
\end{enumerate}

\textbf{HUMAN EXPERT EXAMPLES.} Learn from these actual expert evaluations:

\textit{Example 1 -- Score: 1 (Very Poor).} Question: ``Which expression correctly calculates the sensitivity metric?'' Human Feedback: ``The definition of TN is not given'' $\rightarrow$ Missing variable definitions make question unsolvable.

\textit{Example 2 -- Score: 3 (Acceptable).} Question: ``Which performance metric should replace [MASK]?'' Human Feedback: ``Some variables in choices are not given'' $\rightarrow$ Minor gaps but workable with assumptions.

\textit{Example 3 -- Score: 5 (Excellent).} Question: Complete differential privacy equation with full context. Human Feedback: ``Well-structured with complete information'' $\rightarrow$ Ready for immediate use.

\textbf{CRITICAL EVALUATION GUIDELINES.} Be especially strict about:
\begin{itemize}
    \item \textbf{Missing Variable Definitions:} Any undefined variables $\rightarrow$ Score $\leq$ 2
    \item \textbf{Incomplete Context:} Key background missing $\rightarrow$ Score $\leq$ 2
    \item \textbf{Vague Problem Statements:} Ambiguous questions $\rightarrow$ Score $\leq$ 3
    \item \textbf{Technical Accuracy:} Mathematical/technical errors $\rightarrow$ Score $\leq$ 2
\end{itemize}

\textbf{OUTPUT FORMAT.} Provide assessment in JSON:
\begin{verbatim}
{
  "overall_score": [1-5 integer],
  "dimension_scores": {
    "question_clarity": [1-5],
    "background_relevance": [1-5],
    "answer_accuracy": [1-5],
    "technical_appropriateness": [1-5],
    "mathematical_rigor": [1-5],
    "wireless_relevance": [1-5]
  },
  "binary_flags": {
    "is_correct": [true/false],
    "is_wireless_related": [true/false]
  },
  "quality_analysis": {
    "strengths": ["..."],
    "weaknesses": ["..."],
    "specific_improvements": ["..."]
  }
}
\end{verbatim}

\end{footnotesize}
\end{tcolorbox}

\subsection{Standardized Evaluation Prompts}

To ensure reproducible evaluation, all models receive identical prompts constructed from the following templates:

\begin{tcolorbox}[colback=gray!5,colframe=black!50,title=MCQ Evaluation Template,breakable]
\begin{footnotesize}

\textbf{Background.} [Complete variable definitions and system context]

\textbf{Question.} [Question text]

\textbf{Equation.} [Equation with [MASK] blank]

\textbf{Options.}\\
\textbf{A:} [Option A] \quad \textbf{B:} [Option B] \quad \textbf{C:} [Option C] \quad \textbf{D:} [Option D]

\vspace{0.5em}
\hrule
\vspace{0.5em}

Please analyze this problem step by step. Show your reasoning and calculations.
Your final answer should be given at the end in the format: \texttt{\textbackslash boxed\{X\}} where X is the letter of the correct option.

\end{footnotesize}
\end{tcolorbox}

\begin{tcolorbox}[colback=gray!5,colframe=black!50,title=Fill-in-the-Blank Evaluation Template,breakable]
\begin{footnotesize}

\textbf{Background.} [Complete variable definitions and system context]

\textbf{Question.} [Question text]

\textbf{Equation.} [Equation with [MASK] blank(s)]

\vspace{0.5em}
\hrule
\vspace{0.5em}

Please solve this problem step by step. Fill in the [MASK] blank(s) with the correct mathematical expression(s).

For single mask: Your final answer should be given at the end in the format: \texttt{\textbackslash boxed\{your\_answer\}}.

For multiple masks: Your final answers should be given at the end in the format: \texttt{\textbackslash boxed\{answer1\}, \textbackslash boxed\{answer2\}, \ldots} (for the N blanks in order).

\end{footnotesize}
\end{tcolorbox}

\section{Representative System Model Extractions}
\label{app:model_examples}

This section presents three illustrative system-model extractions from
research papers in the corpus. They are included to show the style and
complexity of source material used for problem generation; they are not
used as item-level validation evidence for the released benchmark.

\subsection{Example 1: Digital Twin-Assisted SIM-Based Air-Ground Communication}

This model integrates multi-layer stacked intelligent metasurface (SIM) beamforming with eVTOL trajectory optimization, representing the convergence of aerial communications and reconfigurable surface technologies.

\begin{tcolorbox}[colback=gray!5,colframe=black!50, title=SIM-Based Air-Ground Communication System,breakable]
\begin{footnotesize}

\paragraph{Background}
This paper proposes a Digital Twin (DT)-assisted framework for joint optimization of Stacked Intelligent Metasurface (SIM)-based air-ground communication and electric Vertical Take-off and Landing (eVTOL) flight control within prescribed air corridors. The system model integrates a multi-layer SIM beamforming structure at the Air Traffic Control (ATCo) station with a composite potential field method for eVTOL trajectory planning, aiming to maximize the sum transmission rate while ensuring safe navigation.

\paragraph{Key Assumptions}
$\bullet$ The air-ground channel between the SIM and each eVTOL follows a Rician fading model.
$\bullet$ Each meta-atom on the SIM imposes an ideal, configurable phase shift without amplitude attenuation.
$\bullet$ The transmission matrices $\mathbf{W}^l$ between metasurface layers are modeled based on Rayleigh-Sommerfeld diffraction theory.
$\bullet$ eVTOLs fly within a predefined, non-overlapping air corridor $\mathcal{R}_{cor}$.
$\bullet$ The signals for different eVTOLs are i.i.d.\ with zero mean and unit variance.
$\bullet$ The additive receiver noise is independent, circularly symmetric complex Gaussian (AWGN).

\paragraph{Parameter Definitions}
$\mathbf{B} = [x^{ATC}, y^{ATC}, 0]^T$ (ATCo station position; $\mathbb{R}^{3 \times 1}$; m); $M$ (eVTOLs / ATCo antennas; $\mathbb{Z}^+$); $L$ (SIM layers; $\mathbb{Z}^+$); $K$ (meta-atoms per layer; $\mathbb{Z}^+$); $N$ (time slots; $\mathbb{Z}^+$); $\delta$ (slot duration; s); $\boldsymbol{q}_m[n] \in \mathcal{R}_{cor} \subset \mathbb{R}^{3}$ (eVTOL position); $V_{max}$ (max velocity; m/s); $P_{ATC}$ (total ATCo power; W); $p_m[n] \in \mathbb{R}^+$ (per-eVTOL power; W); $\theta_k^l[n] \in [0, 2\pi)$ (phase shift; rad); $\boldsymbol{\Psi}^l[n] = \mathrm{diag}(e^{j\theta_1^l[n]}, \ldots, e^{j\theta_K^l[n]})$ (phase-shift matrix; $\mathbb{C}^{K \times K}$); $\mathbf{W}^l \in \mathbb{C}^{K \times K}$ (inter-layer transmission); $\boldsymbol{w}_m^1 \in \mathbb{C}^{K \times 1}$; $\lambda$ (wavelength; m); $d_x, d_y$ (meta-atom size; m); $\boldsymbol{h}_m^H[n] \in \mathbb{C}^{1 \times K}$ (channel vector); $\rho_0$ (reference path loss); $\alpha^h$ (path-loss exponent); $\kappa^h$ (Rician factor; dB); $\sigma_m^2$ (noise power; W); $s_m[n] \in \mathbb{C}$ (data symbol; $\mathbb{E}\{|s_m[n]|^2\}=1$).

\paragraph{Core Equations}

\textbf{1) SIM Beamforming Matrix.}
\begin{align*}
\mathbf{G}[n] = \boldsymbol{\Psi}^L[n] \mathbf{W}^L \boldsymbol{\Psi}^{L-1}[n] \cdots \boldsymbol{\Psi}^2[n] \mathbf{W}^2 \boldsymbol{\Psi}^1[n] \in \mathbb{C}^{K \times K}.
\end{align*}

\textbf{2) Inter-layer Transmission Matrix Entry.}
\begin{align*}
w_{k,k'}^l = \frac{d_x d_y \cos \chi_{k,k'}^l}{d_{k,k'}^l} \left( \frac{1}{2\pi d_{k,k'}^l} - j \frac{1}{\lambda} \right) e^{j 2\pi d_{k,k'}^l / \lambda}.
\end{align*}

\textbf{3) Air-Ground Channel.}
\begin{align*}
h_{m,k}[n] = \sqrt{ \frac{\rho_0}{(d_m[n])^{\alpha^h}} } \sqrt{ \frac{\kappa^h}{\kappa^h + 1} } \bar{h}_m[n].
\end{align*}

\textbf{4) Received Signal.}
\begin{align*}
y_m[n] = \boldsymbol{h}_m^H[n] \mathbf{G}[n] \sum_{m'=1}^{M} \boldsymbol{w}_{m'}^1 \sqrt{p_{m'}[n]} s_{m'}[n] + \tau,\quad \tau \sim \mathcal{CN}(0, \sigma_m^2).
\end{align*}

\textbf{5) SINR.}
\begin{align*}
\mathrm{SINR}_m[n] = \frac{ | \boldsymbol{h}_m^H[n] \mathbf{G}[n] \boldsymbol{w}_m^1 |^2 p_m[n] }{ \sum_{m'\neq m} | \boldsymbol{h}_{m'}^H[n] \mathbf{G}[n] \boldsymbol{w}_{m'}^1 |^2 p_{m'}[n] + \sigma_m^2 }.
\end{align*}

\textbf{6) Achievable Rate.} $R_m[n] = \log\left( 1 + \mathrm{SINR}_m[n] \right)$.

\textbf{7) Joint Optimization (P1).}
\begin{align*}
(\mathrm{P1}):\quad & \max_{\mathbf{P}, \boldsymbol{\Psi}, \mathcal{Q}} \sum_{n=1}^{N} \sum_{m=1}^{M} R_m[n] \\
\text{s.t.}\quad & \mathbf{C1}: \sum_{m=1}^{M} p_m[n] \leq P_{ATC},\ \mathbf{C2}: p_m[n] \geq 0,\ \mathbf{C3}: \theta_k^l[n] \in [0, 2\pi), \\
& \mathbf{C4}: \| \boldsymbol{q}_m[n] - \boldsymbol{q}_m[n-1] \| \leq V_{max} \delta,\ \mathbf{C5}: \boldsymbol{q}_m[n] \in \mathcal{R}_{cor}, \\
& \mathbf{C6}: \boldsymbol{q}_m[0] = \boldsymbol{f}_m[0],\ \boldsymbol{q}_m[N] = \boldsymbol{f}_m[N].
\end{align*}

\textbf{8) Composite Potential Field (CPF) Force.}
$\boldsymbol{a}_i[n] = -\nabla \left( \mathcal{F}_i^{com}[n] + \mathcal{F}_i^{sep}[n] + \mathcal{F}_i^{tar}[n] \right)$,
with field hyperparameters $\{k_{tar}, k_{sep}, k_{com}\}$ optimized via a DQN framework.

\end{footnotesize}
\end{tcolorbox}

\subsection{Example 2: Multi-UAV Patrol Inspection with Mobile Edge Computing}

This system model captures the complexity of joint communication, computation, and trajectory optimization in UAV-enabled MEC networks.

\begin{tcolorbox}[colback=gray!5,colframe=black!50, title=UAV-MEC System Model,breakable]
\begin{footnotesize}

\paragraph{Background}
This paper considers a multi-UAV patrol inspection system where UAVs traverse predetermined cruise points to collect data and offload it to Ground Base Stations (GBSs) equipped with Mobile Edge Computing (MEC) servers. The system model jointly optimizes cruise point assignment, communication scheduling, computational allocation, and UAV trajectory to minimize total energy consumption and balance task completion times among UAVs.

\paragraph{Key Assumptions}
$\bullet$ UAVs fly at a constant altitude $H_U$.
$\bullet$ GBSs are deployed with sufficient density to ensure continuous cellular coverage.
$\bullet$ A TDMA scheme is used for UAV-GBS communication.
$\bullet$ The communication rate model incorporates a logistic function of the elevation angle, based on empirical measurements.
$\bullet$ The information causality constraint must be satisfied (processed data $\leq$ received data).
$\bullet$ UAV dynamics follow a rotary-wing energy consumption model.
$\bullet$ The CPU cycles required per bit ($C_U$) are known and depend on the task type.

\paragraph{Parameter Definitions}
$\mathcal{U} = \{u_1, \ldots, u_N\}$ (set of UAVs); $\mathcal{G} = \{g_1, \ldots, g_M\}$ (set of GBSs); $\boldsymbol{S} = \{s_1, \ldots, s_K\}$ (cruise points); $\mathbf{w}_{s_k} \in \mathbb{R}^{2}$ (cruise point coords; m); $\mathbf{w}_{g_m} \in \mathbb{R}^{2}$ (GBS coords; m); $H_U, H_G$ (UAV/GBS altitude; m); $\boldsymbol{\eta}(t) \in \mathbb{R}^{2}$ (UAV horizontal position); $\mathbf{v}(t)$ (UAV velocity, $\|\mathbf{v}\| \leq V_{max}$); $Q_{s_k}$ (data volume at $s_k$; bits); $R_{g_m}(t)$ (rate to GBS; bps); $\tau_{g_m}(t) \in \{0,1\}$ (scheduling indicator); $f_U, f_{g_m}$ (CPU frequencies; Hz); $C_U$ (cycles/bit); $P(t)$ (UAV transmit power; W); $T_i$ (task completion; s); $\vartheta_U$ (effective capacitance; F).

\paragraph{Core Equations}

\textbf{1) Distance and Elevation Angle.}
\begin{align*}
d_{g_m}(t) &= \sqrt{ ( H_U - H_G )^2 + \left\| \boldsymbol{\eta}(t) - \mathbf{w}_{g_m} \right\|^2 }, \\
\theta_{g_m}(t) &\triangleq \frac{180}{\pi} \arctan\left( \frac{H_U - H_G}{ \| \boldsymbol{\eta}(t) - \mathbf{w}_{g_m} \| } \right).
\end{align*}
Domain: $\theta_{g_m}(t) \in (0^\circ, 90^\circ]$, $d_{g_m}(t) > 0$.

\textbf{2) Communication Rate Model.}
\begin{align*}
R_{g_m}(t) = \left( \chi_3 + \frac{\chi_4}{1 + e^{-(\chi_1 + \chi_2 \theta_{g_m}(t))}} \right) H \log_2 \left(1 + \frac{\hat{\gamma} P(t)}{(d_{g_m}(t))^{\alpha}} \right),
\end{align*}
with $\chi_1<0, \chi_2>0, \chi_4>0, \chi_3+\chi_4=1$; $H$ bandwidth (Hz); $\hat{\gamma} = \beta_0 / (\sigma^2 \Lambda)$.

\textbf{3) Information-Causal Constraint.}
\begin{align*}
\int_0^{T_P} \frac{f_{g_m}(t)}{C_U}\, dt \leq \int_0^{T_P} \tau_{g_m}(t) R_{g_m}(t)\, dt,\quad \forall T_P \in [0, T_i].
\end{align*}

\textbf{4) Energy Consumption Models.}
\begin{align*}
E_t &= \sum_{m=1}^{M} \sum_{k=1}^{K_i} \int_0^{T_{s_k}} \tau_{g_m}(t) P(t)\, dt, \\
E_c &= \int_0^{T_i} \vartheta_U f_U^3(t)\, dt, \\
E_f &= \int_0^{T_i} \Big[ P_0 \big(1 + \tfrac{3 \|\mathbf{v}(t)\|^2}{U_{tip}^2} \big) + P_i \big( \sqrt{1 + \tfrac{\|\mathbf{v}(t)\|^4}{4v_0^4}} - \tfrac{\|\mathbf{v}(t)\|^2}{2v_0^2} \big)^{1/2} \\
& \qquad\qquad + \tfrac{1}{2} d_0 \rho s \hat{a} \|\mathbf{v}(t)\|^3 \Big] dt, \\
E_i &= E_c + E_t + E_f.
\end{align*}

\textbf{5) Original Optimization Problem (P0).}
\begin{align*}
\mathrm{(P0):}\quad \min \sum_{i=1}^{N} \big( E_i + \phi T_i + \lambda (T_i - T_{avg}) \big)
\end{align*}
subject to scheduling, information-causal, data-processing, trajectory, and velocity constraints (decomposed into Task Assignment and Path Planning subproblems).

\end{footnotesize}
\end{tcolorbox}

\subsection{Example 3: RIS-Aided Unsourced Random Access}

\begin{tcolorbox}[colback=gray!5,colframe=black!50,title=RIS-Aided URA System,breakable]
\begin{footnotesize}

\paragraph{Background}
The paper proposes a RIS-aided unsourced random access (URA) system where a massive number of users communicate with a base station (BS) via a reconfigurable intelligent surface (RIS). The direct user-BS links are assumed completely blocked, making the RIS essential for connectivity. The system employs a slotted transmission structure with joint pilot detection, channel estimation, and RIS phase shift optimization.

\paragraph{Key Assumptions}
\begin{itemize}
    \item Quasi-static block fading channels (constant over a frame)
    \item Perfect knowledge of RIS-BS channel $\mathbf{G}$ (stationary elements)
    \item Passive RIS with unit-modulus phase shifts: $|[\mathbf{w}_t]_i| = 1$
    \item Blocked direct user-BS links (no direct path)
    \item Saleh-Valenzuela channel model for RIS-BS and user-RIS links
    \item UPA antenna arrays at both BS and RIS
\end{itemize}

\paragraph{Parameter Definitions}
$\mathbf{G} \in \mathbb{C}^{M \times N}$ (RIS-BS channel); $\mathbf{h}_i \in \mathbb{C}^{1 \times N}$ (user-RIS channel); $\mathbf{w}_t \in \mathbb{C}^{N \times 1}$ (RIS phase shift, $|[\mathbf{w}_t]_i| = 1$); $x_{i,t} \in \mathbb{C}$ (transmitted symbol); $\mathbf{z}_t \sim \mathcal{CN}(\mathbf{0}, \sigma_z^2\mathbf{I}_M)$; $M$ (BS antennas); $N$ (RIS elements); $K_a$ (active users); $n$ (channel uses); $L_G, L_{R,i}$ (paths in RIS-BS / user-RIS channels).

\paragraph{Core Equations}
\textbf{1) Received Signal Model.}
\begin{align*}
\mathbf{y}_t = \sum_{i=1}^{K_a} \mathbf{G}\, \mathrm{diag}(\mathbf{h}_i)\, \mathbf{w}_t\, x_{i,t} + \mathbf{z}_t,\quad t = 1,\ldots,n.
\end{align*}

\textbf{2) Pilot Phase Received Signal.}
\begin{align*}
\mathbf{Y}_p = \sqrt{P_p} \sum_{i \in \mathcal{S}_s} \mathbf{G}\, \mathrm{diag}(\mathbf{h}_i)\, \mathbf{W}_{p_s}\, \mathrm{diag}(\mathbf{p}_i) + \mathbf{Z}_p.
\end{align*}

\textbf{3) Data Phase Received Signal.}
\begin{align*}
\mathbf{Y}_{c,f} = \sqrt{P_c} \sum_{i \in \mathcal{S}_s} \mathbf{G}\, \mathrm{diag}(\mathbf{h}_i)\, \mathbf{W}_{c_s}\, \mathrm{diag}(\mathbf{b}_i)\, v_{i,f} + \mathbf{Z}_{c,f},\quad v_{i,f} \in \{\pm 1\}.
\end{align*}

\textbf{4) Channel Model -- RIS-BS.}
\begin{align*}
\mathbf{G} = \sqrt{MN} \sum_{l=1}^{L_G} \mu_l\, \mathbf{a}_M(\phi_{r,l}, \psi_{r,l})^T\, \mathbf{a}_N(\phi_{t,l}, \psi_{t,l}),\quad \mu_l \sim \mathcal{CN}(0, L_0 d_l^{-\alpha_{PL}}).
\end{align*}

\textbf{5) Channel Model -- User-RIS.}
\begin{align*}
\mathbf{h}_i = \sqrt{N} \sum_{f_i=1}^{L_{R,i}} \mu_{f_i}\, \mathbf{a}_N(\phi_{i,f_i}, \psi_{i,f_i}),\quad \mu_{f_i} \sim \mathcal{CN}(0, L_0 d_{f_i}^{-\alpha_{PL}}).
\end{align*}

\textbf{6) Steering Vector.}
\begin{align*}
\mathbf{a}_N(\phi, \psi) = \frac{1}{\sqrt{N}} e^{-j2\pi\bar{\phi}\mathbf{n}_1} \otimes e^{-j2\pi\bar{\psi}\mathbf{n}_2},
\end{align*}
with $\bar{\phi} = \sin(\phi)\cos(\psi)$, $\bar{\psi} = \sin(\psi)$, antenna spacing $d = \lambda/2$.

\paragraph{Illustrative Checks}
\begin{itemize}
    \item \textit{Dimensional consistency:} all matrix multiplications consistent (e.g., $\mathbf{G} \in \mathbb{C}^{M \times N}$ times $\mathrm{diag}(\mathbf{h}_i) \in \mathbb{C}^{N \times N}$ yields $\mathbb{C}^{M \times N}$).
    \item \textit{Boundary conditions:} unit-modulus constraint $|[\mathbf{w}_t]_i| = 1$ enforced for passive RIS.
    \item \textit{Parameter homogeneity:} channel gains $\mu_l$, $\mu_{f_i}$ have consistent units.
    \item \textit{Physical constraints:} angle parameters restricted to $[0,2\pi)$, steering vectors normalized.
\end{itemize}

\end{footnotesize}
\end{tcolorbox}

\subsection{Use of These Examples}

These examples are qualitative illustrations of the extraction format:
variable definitions, assumptions, equations, and constraints are kept
together so that downstream problem generation can reference them. The
released benchmark is validated at the problem level through the QA
pipeline of Section~\ref{sec:qa}; these appendix examples are not an
independent technical audit of the source papers or the extraction
model.

\section{Human Quality-Assessment Rubric}
\label{app:quality_rubric_human}

To structure the manual verification process, we used a standardized
five-point assessment rubric. This rubric guides domain experts to
evaluate generated questions based on mathematical correctness, problem
clarity, and domain relevance. The explicit criteria make the review
process auditable, but the raw agreement estimate is low and is
reported as a limitation in Appendix~\ref{app:expert_agreement}. The detailed
definitions for each score level are presented in
Table~\ref{tab:quality_rubric_detailed}.

\begin{table}[t]
\centering
\caption{Detailed expert question quality assessment rubric}
\label{tab:quality_rubric_detailed}
\scriptsize
\begin{tabular}{p{0.13\textwidth}p{0.78\textwidth}}
\toprule
\textbf{Score} & \textbf{Criteria} \\
\midrule
\textbf{1 - Invalid} &
Problem statement or solution is clearly wrong or contradictory; Not related to wireless communications domain; Cannot be used as a valid question \\
\midrule
\textbf{2 - Poor} &
Statement correct but problem too trivial (answerable instantly); Problem too vague or nearly impossible to answer correctly; Very little learning or evaluation value \\
\midrule
\textbf{3 - Acceptable} &
Statement and solution reasonable with no major errors; Difficulty and relevance are average; Can be kept but adds limited value (baseline quality) \\
\midrule
\textbf{4 - Good} &
Clear and well-structured problem; Relevant to domain and moderately challenging; Provides meaningful assessment of understanding; Worth keeping and recommending \\
\midrule
\textbf{5 - Excellent} &
Highly relevant to the domain; Strong depth, creativity, or insight required; Excellent for differentiating levels of understanding; Strongly recommended for inclusion \\
\bottomrule
\end{tabular}
\end{table}

\section{Large Language Model-Assisted Quality Assessment}
\label{app:llm_quality_assessment}

LLM-assisted assessment is a filtering stage, not the final quality authority.
It removes obvious extraction failures before domain-specialist review and
provides a structured score trace for auditability. Final release membership is
determined after expert validation, as described in Section~\ref{sec:qa} and
Appendix~\ref{app:expert_agreement}.

\subsection{Quality Assessment Framework}

The rubric evaluates each candidate problem along six dimensions:

\begin{enumerate}
    \item \textbf{Question clarity} (1--5): whether the requested output is unambiguous.
    \item \textbf{Background relevance} (1--5): whether the provided context is sufficient and on topic.
    \item \textbf{Answer accuracy} (1--5): whether the target answer is mathematically valid and well formatted.
    \item \textbf{Technical appropriateness} (1--5): whether the difficulty matches the intended benchmark scope.
    \item \textbf{Mathematical rigor} (1--5): whether notation, dimensions, and conventions are coherent.
    \item \textbf{Wireless relevance} (1--5): whether the problem depends on wireless-communications structure.
\end{enumerate}

\subsection{Illustrative Annotation Checks}

Table~\ref{tab:annotation_examples} shows representative score traces. The
examples are included to expose the filter's behaviour, including cases where
the LLM score is too optimistic; they are not evidence that the LLM assessor is
an expert substitute.

\begin{table}[ht]
\centering
\caption{Illustrative LLM-assisted annotation checks across quality levels.}
\label{tab:annotation_examples}
\scriptsize
\resizebox{.95\linewidth}{!}{
\begin{tabular}{p{0.18\textwidth}p{0.38\textwidth}p{0.38\textwidth}}
\toprule
\textbf{Quality Level} & \textbf{Question Content} & \textbf{LLM Assessment} \\
\midrule
\textbf{High Quality} &
\textbf{Background:} Federated fine-tuning system with low-rank adaptation matrices $\mathbf{A}_k \in \mathbb{R}^{d \times r}$, $\mathbf{B}_k \in \mathbb{R}^{r \times d}$\par
\textbf{Question:} Which term completes: $\mathbf{W} + [\text{MASK}]$?\par
\textbf{Options:} A) $\mathbf{A}_k\mathbf{B}_k$ B) $\mathbf{B}_k\mathbf{A}_k$\par
\textbf{Human:} 4/5 &
\textbf{LLM Score:} 4/5\par
\textbf{Strengths:} Clear structure, complete context, accurate answer, rigorous notation\par
\textbf{Weaknesses:} Could benefit from brief explanation of low-rank adaptation significance\par
\textbf{Agreement:} $\checkmark$ \\
\midrule
\textbf{Medium Quality} &
\textbf{Background:} H-NOMA system with variable definitions partially provided\par
\textbf{Question:} Fill in [MASK] for the equation\par
\textbf{Human:} 2/5 &
\textbf{LLM Score:} 3/5\par
\textbf{Strengths:} Wireless relevance, accurate answer\par
\textbf{Weaknesses:} Ambiguous [MASK] usage, lacks clarity in instructions\par
\textbf{Technical Issues:} Missing variable definitions\par
\textbf{Disagreement:} LLM too optimistic \\
\midrule
\textbf{Low Quality} &
\textbf{Background:} Transformer model context with incomplete variable definitions\par
\textbf{Question:} What replaces the ``full key matrix''?\par
\textbf{Human:} 1/5 &
\textbf{LLM Score:} 3/5\par
\textbf{Strengths:} Clear structure, accurate answer\par
\textbf{Weaknesses:} Limited wireless relevance, focuses more on tensor parallelism\par
\textbf{Technical Issues:} ``Full key matrix'' not defined\par
\textbf{Bias:} LLM shows optimistic scoring pattern \\
\bottomrule
\end{tabular}
}
\end{table}

\section{Expert Screening Agreement and Examples}
\label{app:expert_examples}

This section first reports the retained expert-screening coverage and
agreement analysis, then presents representative examples from the
expert evaluation process across different score levels.

\subsection{Expert-Screening Records and Agreement}
\label{app:expert_agreement}

The retained expert-screening records contain $9{,}222$ non-empty rows
from five PhD-student or postdoctoral domain specialists, covering
$5{,}205$ unique candidate question IDs. All $4{,}027$ released
problem ids appear in these records, giving a historical release
acceptance rate of $77.4\%$ ($4{,}027/5{,}205$). The records include blank scores for $1{,}496$ rows; these
rows are retained as provenance but excluded from agreement statistics.

For the retrospective agreement analysis, we collapse duplicate
same-rater rows for the same question by majority vote over non-empty
scores, using the later CSV row only for tied duplicate majorities. This
cleaning affects $168$ duplicate question-rater pairs, including $74$
with conflicting non-empty scores. After cleaning, $2{,}454$ candidate
questions have at least two scored raters, yielding $2{,}456$ pairwise
rater observations. Pooled pairwise exact agreement on the five-class
rubric is $32.5\%$; unweighted five-class Cohen's $\kappa$ is $0.101$;
quadratic-weighted five-class $\kappa$ is $0.211$ with a problem-level
bootstrap $95\%$ CI of $[0.169,0.252]$. For the binary inclusion decision
(score $\geq 3$ vs. $<3$), agreement is $64.1\%$ and Cohen's $\kappa$ is
$0.164$ with bootstrap $95\%$ CI $[0.124,0.205]$.

The disagreement is concentrated near the screening boundary. Although
exact five-class agreement is $32.5\%$, $79.6\%$ of pairs are within one
rubric point. Of the $882$ binary include/exclude disagreements, $45.2\%$
are exactly the $2$-versus-$3$ pair. The pooled include prevalence is
$69.2\%$, which contributes to the gap between raw agreement and
$\kappa$ but does not rescue the low reliability.

These values should be read conservatively. They document the historical
screening process and support the $77.4\%$ acceptance-rate provenance, but
they do not establish high inter-rater reliability, explain why individual
raters disagreed, or show how revision changed error rates. We therefore use
the archive as screening-process provenance rather than as a strong reliability
or error-reduction claim. The verifier robustness analysis in
Appendix~\ref{app:evaluation_robustness_app} addresses a separate
evaluation-stage risk and does not repair this construction-stage
agreement limitation.

\paragraph{Selection rule for the examples.} Each example under Scores 5, 4, and 3 is a released problem on which every recorded expert score equals that score. Below-threshold examples are candidates that every rater scored 1 or 2; they were excluded during screening and are not part of the release.

\subsection{Score 5 -- Excellent Quality}

Questions scoring 5 demonstrate comprehensive variable definitions, clear mathematical structure, and strong pedagogical value.

\begin{questionbox}[Question ID: 1484]
\textbf{Paper:} \href{https://arxiv.org/pdf/2508.11971v1}{2508.11971v1}

\textbf{Answer:} $\alpha_0 \leq \gamma \dfrac{(d_1 + \sqrt{2}\epsilon/2)^2}{(d_1 - \sqrt{2}\epsilon/2)^2}$.

\textbf{Background.} In the area-discretization approach to mobile-charger positioning, the approximation ratio bounds the power variation within a grid. $\alpha_0$ approximation ratio; $\gamma$ beam power reduction factor (typically $1.25$--$2.5$); $d_1$ distance from the charger to the grid center (m); $\epsilon$ edge length of the square grid (m).

\textbf{Question.} Write the complete inequality for the approximation ratio bound.
\end{questionbox}

\begin{questionbox}[Question ID: 4129]
\textbf{Paper:} \href{https://arxiv.org/pdf/2506.23817v1}{2506.23817v1}

\textbf{Answer:} $\widetilde{\delta}_t(\Upsilon_t) = \rho \sqrt{ \dfrac{ \Upsilon_t^2 - \Upsilon_{\mathrm{min}}^2 }{ \left( \frac{h}{r_{\mathrm{e}} + h} \right)^2 + \left( \frac{r_{\mathrm{e}}}{r_{\mathrm{e}} + h} \right) \Upsilon_t^2 } }$.

\textbf{Background.} For small central angles in a LEO satellite channel, the exact Doppler shift is approximated with $\cos(x) \approx 1 - x^2/2$ and $\sin(x) \approx x$. $\widetilde{\delta}_t$ approximated Doppler shift (Hz); $\rho$ normalization constant (Hz); $\Upsilon_t$ central angle at time $t$ (rad); $\Upsilon_{\mathrm{min}}$ minimum central angle (rad); $r_{\mathrm{e}}$ Earth's radius (m); $h$ satellite altitude (m).

\textbf{Question.} Write the complete small-angle approximation for the Doppler shift.
\end{questionbox}

\begin{questionbox}[Question ID: 3719]
\textbf{Paper:} \href{https://arxiv.org/pdf/2508.12964v1}{2508.12964v1}

\textbf{Answer:} $y(n\tau T) = \sum_{l=0}^{L-1} \sum_k h_l a_k x((n-k-\delta_l)\tau T) + w(n\tau T)$.

\textbf{Background.} In a quasi-static multipath Rayleigh fading channel, the faster-than-Nyquist (FTN) received signal combines intentional ISI with channel-induced effects. $y(n\tau T)$ received signal; $h_l$ complex coefficient of the $l$-th path; $\delta_l$ delay of the $l$-th path (symbols); $a_k$ transmitted symbol; $x(t) = g(t) * g(-t)$ ISI kernel; $w(t)$ AWGN; $L$ number of multipath components.

\textbf{Question.} Write the complete received signal equation for the multipath FTN channel.
\end{questionbox}

\subsection{Score 4 -- Good Quality}

\begin{questionbox}[Question ID: 13059]
\textbf{Paper:} \href{https://arxiv.org/pdf/2012.07034v1}{2012.07034v1}

\textbf{Answer:} $\hat{\gamma}_w = \dfrac{(1 - \alpha_s) \gamma_w}{1 + \alpha_s \gamma_w}$.

\textbf{Background.} In a downlink NOMA pair, the weak user $U_w$ decodes its signal by treating the strong user's signal as interference, without SIC. $\hat{\gamma}_w$ effective NOMA SINR of $U_w$; $\alpha_s$ power allocation factor of the strong user; $\gamma_w = P_t |h_w|^2 / (N_0 + I)$ OMA SINR of $U_w$.

\textbf{Question.} Write the complete equation for the NOMA weak user's SINR.
\end{questionbox}

\begin{questionbox}[Question ID: 12806]
\textbf{Paper:} \href{https://arxiv.org/pdf/2407.09944v2}{2407.09944v2}

\textbf{Answer:} $(\varepsilon_{eff} - 0.258)\left(\frac{W}{h} + 0.8\right)$.

\textbf{Background.} In microstrip patch antenna design, fringing fields extend the electrical length of the patch beyond its physical length: $\Delta L = 0.412h \frac{(\varepsilon_{eff} + 0.3)\left(\frac{W}{h} + 0.264\right)}{[\text{MASK}]}$, with $h$ substrate thickness (m), $\varepsilon_{eff}$ effective dielectric constant, and $W$ patch width (m).

\textbf{Question.} What is the denominator term in the length extension formula?
\end{questionbox}

\subsection{Score 3 -- Acceptable Quality}

\begin{questionbox}[Question ID: 11710]
\textbf{Paper:} \href{https://arxiv.org/pdf/2403.11806v1}{2403.11806v1} \quad \textbf{Answer:} B.

\textbf{Background.} A base station separates $N$ users with a zero-forcing (ZF) combiner. $\mathbf{w}_n \in \mathbb{C}^{M \times 1}$ ZF combining vector of user $n$; $p_n$ transmit power of user $n$ (W); $\sigma^2$ noise power at the BS (W); $R_n$ achievable rate of user $n$.

\textbf{Question.} Which expression correctly represents the argument of the logarithm in $R_n = \log_2 \left( 1 + [\text{MASK}] \right)$ for a ZF receiver?

\textbf{Options.}
A: drops the combiner norm, $p_n / \sigma^2$;
B: $\dfrac{p_n}{\| \mathbf{w}_n \|_2^2 \sigma^2}$ (correct);
C: places $\| \mathbf{w}_n \|_2^2$ in the numerator;
D: uses $\| \mathbf{w}_n \|_2^4$.
\end{questionbox}

\begin{questionbox}[Question ID: 1595]
\textbf{Paper:} \href{https://arxiv.org/pdf/2503.02156v1}{2503.02156v1} \quad \textbf{Answer:} C.

\textbf{Background.} An RF fingerprinting framework trains a CNN on standardized spectrograms with a triplet loss. $f(x_i) \in \mathbb{R}^{D}$ L2-normalized embedding of spectrogram $x_i$; $x_i$, $x_j$, $x_k$ anchor, positive, and negative samples; $\alpha_{\mathrm{triplet}}$ margin.

\textbf{Question.} Which expression is the core term inside the hinge function in $L_{\mathrm{triplet}} = \sum_{i,j,k}^{N} \left[ [\text{MASK}] \right]_+$?

\textbf{Options.}
A: swaps the positive and negative distances;
B: subtracts the margin;
C: $\| f(x_i) - f(x_j) \|_2^2 - \| f(x_i) - f(x_k) \|_2^2 + \alpha_{\mathrm{triplet}}$ (correct);
D: swaps the distances and subtracts the margin.
\end{questionbox}

\subsection{Below Threshold -- Excluded During Screening}

In these examples, the masked content is a relation with no mathematical form or a bare operator.

\begin{questionbox}[Question ID: 658]
\textbf{Paper:} \href{https://arxiv.org/pdf/2505.02076v1}{2505.02076v1} \quad \textbf{Answer:} $\text{Computational Cost}$, $N_{\text{tokens}}$.

\textbf{Background.} The computational cost of operating an LLM agent is related to $N_{\text{tokens}}$, the total number of tokens the model consumes (thousands): $[\text{MASK}] \propto [\text{MASK}]$.

\textbf{Question.} Complete the relationship for the system's computational cost.
\end{questionbox}

\begin{questionbox}[Question ID: 12563]
\textbf{Paper:} \href{https://arxiv.org/pdf/2502.16592v1}{2502.16592v1} \quad \textbf{Answer:} \texttt{\textbackslash frac\{\}} and $/$.

\textbf{Background.} In antenna pattern analysis, the power pattern is normalized by its maximum. $\mathbf{U}(\theta, \phi)$ normalized power pattern; $|\mathbf{E}(\theta, \phi)|^2$ power density; $\max(|\mathbf{E}(\theta, \phi)|^2)$ its maximum over all angles: $\mathbf{U}(\theta, \phi) = [\text{MASK}]\ ( | \mathbf{E}(\theta, \phi) |^2 )\ [\text{MASK}]\ ( \max( | \mathbf{E}(\theta, \phi) |^2 ) )$.

\textbf{Question.} Complete the operator and the functional relationship.
\end{questionbox}

\begin{questionbox}[Question ID: 7739]
\textbf{Paper:} \href{https://arxiv.org/pdf/2401.03333v1}{2401.03333v1}

\textbf{Answer:} $\mathrm{MIL} = f(\mathrm{WUR\ Type},\ \mathrm{Modulation},\ \mathrm{Payload\ Size},\ \mathrm{Duration},\ \mathrm{Bandwidth})$.

\textbf{Background.} In a 5G NR IoT system with wake-up receivers (WURs), the maximum isotropic loss is the largest path loss a wake-up signal (WUS) can endure while still being detected with a target probability. $\mathrm{MIL}$ maximum isotropic loss (dB); $\mathrm{WUR\ Type}$ receiver architecture (OFDM- or OOK-based); $\mathrm{Modulation}$, $\mathrm{Payload\ Size}$ (bits), $\mathrm{Duration}$ (s), $\mathrm{Bandwidth}$ (Hz) of the WUS.

\textbf{Question.} Write the complete functional relationship for the Maximum Isotropic Loss.
\end{questionbox}

\section{Datasheet for Datasets}
\label{app:datasheet}

This datasheet documents WirelessMathBench-XL following the Datasheets for
Datasets framework~\citep{gebru2021datasheets}. The dataset is hosted at
\url{https://huggingface.co/datasets/XINLI1997/WirelessMATHBench-XL}, code at \url{https://github.com/LiXin97/WirelessMathBench-XL}, and the project page at \url{https://lixin.ai/WirelessMathBench-XL}.

\subsection{Motivation}

\paragraph{For what purpose was the dataset created?}
WirelessMathBench-XL was created as a training-scale benchmark for
\emph{implicit-constraint mathematical reasoning} as instantiated in
wireless-communications research, and as an \emph{auditable}
evaluation resource with one fixed prompt-surface lexical-overlap channel. The
dataset is intended to support verification-based reinforcement learning
on constraint-grounded mathematics, and to provide a reusable instrument
for measuring prompt-surface lexical overlap between benchmark problems
and auditable public pretraining slices.

\paragraph{Who created the dataset and on behalf of which entity?}
Created by the paper authors at Nanyang Technological University.

\paragraph{Who funded the creation?}
Gemini API credits provided by Google through the Gemini Academic Program.

\subsection{Composition}

\paragraph{What do the instances represent?}
Each instance is a self-contained mathematical problem derived from a
single wireless-communications arXiv paper. An instance bundles:
problem text, variable definitions with units and domain restrictions,
underlying assumptions, boundary conditions, the verifiable
ground-truth expression, the source-paper identifier, the task format
(MCQ / Fill-in-25/50/75 / FEC), and
the contamination metadata (per-problem $13$-gram hit count
against RedPajama-arXiv, $S_0$/$S_1$ membership flags, and a capped
list of matched $n$-grams for spot inspection on distributed records). Technical subfield
labels are reported as aggregate corpus statistics in this release, not as
instance-level release fields.

\paragraph{How many instances are there in total?}
$4{,}027$ problems from $836$ retained arXiv source papers selected
from a $970$-paper source pool spanning July $2005$--August $2025$,
partitioned into a problem-level release split of
$3{,}227$ training and $800$ test problems. The release split is not
source-paper-disjoint: $456$ retained papers have problems in both
partitions; $34$ test problems from $29$ papers are disjoint from the
training-paper pool.

\paragraph{Does the dataset contain all possible instances or a sample?}
A filtered sample. Stage one is a deterministic relevance score over
candidate arXiv papers across $24$ categories; stage
two is a GPT-4o verifier selecting candidates with
substantial mathematical content; from these, a $970$-paper source pool
enters extraction, and $836$ retained papers contribute accepted
problems after QA (Section~\ref{sec:collection}).
Coverage is by relevance, not by exhaustive enumeration.
The corpus is wireless-centered rather than strictly
physical-layer-only: a small number of adjacent networked-systems,
signal-processing, and edge-computing records passed the relevance and
QA filters and remain in the release. We keep them because they exercise
the same mathematical constraint patterns, but users needing a narrower
wireless-only slice should filter by source paper or future subfield
labels.

\paragraph{What data does each instance consist of?}
Structured JSONL records. Fields: \texttt{problem\_id},
\texttt{paper\_id}, \texttt{paper\_title}, \texttt{type},
\texttt{split}, \texttt{background}, \texttt{question\_text},
\texttt{equation}, \texttt{options} (MCQ only),
\texttt{correct\_answer}, \texttt{prompt}, a compatibility
\texttt{text} field, and a \texttt{contamination} block containing
\texttt{hits\_total}, \texttt{docs\_total}, \texttt{in\_S0},
\texttt{in\_S1}, and \texttt{matched\_ngrams\_capped}. The
\texttt{equation} field is empty for a small number of natural-language
MCQ/FEC items whose prompt is fully specified by \texttt{background} and
\texttt{question\_text}; \texttt{question\_text} and
\texttt{correct\_answer} are populated for every one of the $4{,}027$ records.
The Hugging Face default configurations (\texttt{cc\_by}, \texttt{cc\_by\_sa},
\texttt{cc\_by\_nc\_sa}) expose the same distributed records as Parquet with a
flat schema: \texttt{question\_id} in place of \texttt{problem\_id}; flat
\texttt{in\_S0}, \texttt{in\_S1}, \texttt{hits\_total}, and
\texttt{docs\_total} columns; and no \texttt{split}, \texttt{paper\_title},
\texttt{text}, or \texttt{matched\_ngrams\_capped} fields. In the JSONL
files \texttt{paper\_title} is a placeholder; \texttt{paper\_id} identifies
the source paper.

\paragraph{Is there a label or target associated with each instance?}
Yes---each instance has a verifiable ground-truth answer. The
evaluator (Section~\ref{sec:qa}) checks predictions against this target with
a GPT-4.1-mini semantic judge; a rule-based exact-format scorer runs as a
separate lower-bound check. An early cross-judge probe on a two-stage grader
measured $3.4$--$9.3\%$ absolute per-model worst-case variation
(Appendix~\ref{app:evaluation_robustness_app}).

\paragraph{Is any information missing from individual instances?}
No core field is missing for solving or evaluating the released tasks.
Constraint-class taxonomy labels from Section~\ref{sec:taxonomy_classes} and
technical subfield labels are not instance-level release fields; they are documented as descriptive corpus constructs rather than
supervised label sets.

\paragraph{Are relationships between individual instances made explicit?}
Yes. \texttt{paper\_id} groups instances by source paper. It is a
clustering key, not the unit of the release split: source papers
can contribute problems to both train and test. Finer-grained extraction
record or problem-family identifiers are not shipped; users
needing that grouping should treat all problems from the same
\texttt{paper\_id} as the conservative cluster.

\paragraph{Are there recommended data splits?}
Yes: the problem-level $80/20$ train/test split shipped with the
release. It is the split used for the reference calibration table and
WirelessMathLM training runs, but it should not be interpreted as a
paper-disjoint generalisation split. Users requiring source-paper
disjointness should construct a new split by grouping on
\texttt{paper\_id}; the shipped test side contains only $34$ items from
papers absent from the training pool. The fixed $13$-gram
prompt-surface lexical-overlap audit channel against RedPajama-ArXiv defines
two filtered views: $\mathcal{S}_0$ ($3{,}853$ corpus records, $770$ test
records; $\texttt{hits\_total}=0$) and $\mathcal{S}_1$ ($3{,}985$
corpus records, $792$ test records; $\le 2$ distinct RedPajama documents).
We recommend reporting $\mathcal{S}_0$ as the strict zero-hit view and
$\mathcal{S}_1$ as a lenient sensitivity view, not as general cleanliness
labels (Section~\ref{sec:contamination-audit}). The $\mathcal{S}_1$
count uses the doc-unique convention adopted in the final audit. The
headline subset $\mathcal{S}_0$ is unaffected by this convention.

\paragraph{Are there errors, sources of noise, or redundancies?}
Yes, and they are bounded where measured. (a) Verifier
variation: an early cross-judge probe on a two-stage grader measured
$3.4$--$9.3\%$ absolute per model, characterised in
Appendix~\ref{app:evaluation_robustness_app}; this is neither a bound on the
locked tables nor a direct human-ground-truth false-negative estimate.
(b) Expert-screening agreement: the retained five-rater records support
the $77.4\%$ acceptance-rate provenance but yield low retrospective
agreement (binary inclusion $\kappa=0.164$, bootstrap $95\%$ CI
$[0.124,0.205]$; Appendix~\ref{app:expert_agreement}). We therefore do not
claim high inter-rater reliability for this release. (c) Extraction noise:
DeepSeek-R1 may misparse
non-standard notation in older 3G/4G-era papers. The Tier-2 expert pass
screened these candidates, and the $77.4\%$ acceptance rate records the
fraction retained after QA; the archived scores do not establish how
many extraction errors screening removed.

\paragraph{Is the dataset self-contained, or does it link to or rely on external resources?}
Self-contained at the instance level: every problem ships with its
ground-truth expression and complete variable / assumption block.
Source-paper provenance is preserved as \texttt{paper\_id}
pointing into the public arXiv (\url{https://arxiv.org}); the source
papers themselves are not redistributed---only the derived problems
are. The contamination audit further references the public
RedPajama-arXiv slice~\citep{weber2024redpajama} as the audit
reference; the audit \emph{outputs} (per-problem hit counts for all problems,
\texttt{matched\_ngrams\_capped} for distributed records) are bundled with the
dataset, but the
RedPajama corpus itself is not redistributed. The audited
RedPajama-ArXiv snapshot contains $1{,}558{,}306$ records across $100$
shards, with a latest recorded timestamp of 7 March $2023$. Source
papers dated April $2023$ or later therefore cannot be flagged through
their own source text in this snapshot.

\paragraph{Does the dataset contain confidential, offensive, or personally identifiable data?}
No. Source papers are public arXiv preprints; problems are
mathematical derivations with no personal data. No demographic data,
no faces, no audio, no health records.

\subsection{Collection Process}

\paragraph{How was the data acquired?}
Source-paper acquisition: arXiv API queries across $24$ categories
(core wireless: \texttt{cs.NI}, \texttt{eess.SP}, \texttt{cs.IT};
ML/AI: \texttt{cs.LG}, \texttt{stat.ML}; adjacent fields), filtered by
the deterministic relevance score of Section~\ref{sec:collection} and a
GPT-4o second-stage verifier. Problem generation: DeepSeek-R1
extraction over each paper's LaTeX source (Appendix~\ref{app:llm_disclosure}).

\paragraph{What mechanisms or procedures were used?}
Custom Python collection / filtering pipeline using arXiv's bulk
metadata API; LaTeX source obtained from arXiv's e-print service.
Extraction runs through DeepSeek-R1 via API; rubric scoring uses
GPT-4o; the final contamination audit uses a custom reverse-probe
$n$-gram pipeline over local mirrors of the RedPajama-arXiv slice.

\paragraph{If the data is a sample, what was the sampling strategy?}
Deterministic-score-then-LLM-verifier filter, biased toward recent and
mathematically-rigorous papers. Coverage is by relevance, not
exhaustive; the temporal skew (5G-Advanced/6G-era papers form $65.9\%$
of the corpus) is intentional and documented in Section~\ref{sec:stats}.

\paragraph{Who was involved and how were they compensated?}
Five PhD-student or postdoctoral domain specialists conducted Tier-2
expert validation, with expertise across optimisation theory,
information theory, signal processing, and network analysis. All were
paper authors or institutional research collaborators contributing as
part of salaried research duties; no crowdworkers or external piece-rate
annotators were hired, and no per-task compensation was paid.
The validation rubric and instructions are reproduced in
Appendix~\ref{app:quality_rubric_human}.

\paragraph{Over what timeframe was the data collected?}
Source-paper collection: rolling arXiv harvest throughout 2025, frozen
on September 25, 2025. The harvest covered arXiv submissions in
twenty-four wireless-relevant categories. Source papers themselves span
July $2005$--August $2025$. Problem generation and Tier-2 expert review
ran from 2025 through early 2026. The final contamination audit and
cleaned-subset construction were completed in April 2026.

\paragraph{Were any ethical review processes conducted?}
The dataset is constructed from publicly available arXiv source papers
and contains no personal, behavioural, biometric, or otherwise
human-subject data. The Tier-2 expert review process evaluates
mathematical artifacts (variable definitions, constraints, ground-truth
expressions) as research QA by authors / institutional collaborators,
not as an intervention, survey, or behavioural study of the validators.
Institutional ethics-review approval was therefore not required for the
released dataset.

\paragraph{Did you collect data from individuals directly or via third parties?}
Via third party (arXiv). No individual data subjects.

\subsection{Preprocessing, Cleaning, and Labeling}

\paragraph{Was preprocessing/cleaning/labeling done?}
Yes, in three stages: (a) deterministic + LLM-verifier source-paper
filtering; (b) DeepSeek-R1 extraction into structured records,
followed by Tier-1 GPT-4o rubric scoring and Tier-2 expert validation
(Section~\ref{sec:qa}); (c) final contamination audit producing per-problem
$n$-gram overlap scores against the RedPajama-arXiv slice. The release
records contain only the final validated metadata used by the paper claims.

\paragraph{Was the raw data saved in addition to the cleaned data?}
Yes. Intermediate records are retained: the crawl database and scraper
logs; the selected paper identifiers and scoring weights;
the structured DeepSeek-R1 extraction records; the Tier-1 rubric
scores; and the Tier-2 expert-screening records used to derive the
$77.4\%$ release acceptance rate. The contamination-audit release retains the
final reverse-probe outputs and validation summaries needed to reproduce
$\mathcal{S}_0$/$\mathcal{S}_1$ membership.

\paragraph{Is the preprocessing software available?}
Partly. The code repository contains the contamination-audit pipeline
(reverse-probe $n$-gram audit with \texttt{xxhash64},
$\mathcal{S}_0$/$\mathcal{S}_1$ thresholding, split construction, and
paired-bootstrap evaluation scripts) under the MIT License at \url{https://github.com/LiXin97/WirelessMathBench-XL}. The construction
prompts and rubric are given in Appendices~\ref{app:prompt_construction}
and~\ref{app:quality_rubric_human}; the construction scripts are not
released. The historical
construction path used stochastic LLM calls (DeepSeek-R1 extraction and GPT-4o
filtering and scoring) and is therefore not bitwise reproducible: re-running it
on the same sources yields different problems. Reproducibility claims apply to
the shipped records, their provenance, and the released evaluation results, not to
deterministic regeneration of the corpus.

\subsubsection{Audit implementation note}
\label{app:audit-method-iterations}

The implementation indexes the benchmark side rather than the corpus side.
This direction is required by the scale asymmetry: the benchmark contributes
approximately $3\times10^5$ unique prompt-surface $13$-grams, while the audited
public slice contributes on the order of $10^{10}$ corpus $13$-grams. The
reported labels therefore come from a streaming reverse-probe audit in which
corpus files are scanned against the smaller benchmark index. This design is
the release-defining audit channel for the release; it bounds prompt-surface lexical
overlap against RedPajama-arXiv at the fixed $13$-gram threshold and does not
certify against paraphrase, answer-only, alternate-version, or closed-corpus
exposure.

\subsubsection{N-gram overlap audit method}
\label{app:audit-method}

\textbf{Reverse-probe design.}
We index the small side. From each of the $4{,}027$ benchmark problems
we extract $13$-grams over the concatenation of problem text, variable
definitions, and constraints, after a normalisation pass: lower-case,
strip LaTeX command tokens (\texttt{\textbackslash cmd}, \texttt{\{},
\texttt{\}}, \texttt{\$}), and collapse whitespace. Each $n$-gram is
hashed with \texttt{xxhash64}, and the resulting probe index is a
\texttt{dict} from $n$-gram hash to the list of problem ids that contain
it---approximately $311$K unique entries. Slice
files are streamed in parallel; each worker normalises
and tokenises a file, computes per-line $13$-gram hashes, looks each up
in the probe index, and on a hit increments the per-problem hit counter
for that slice and (capped at a small constant) records the matched
$n$-gram for spot inspection. Workers reduce per-file dictionaries into
a master per-(problem, slice) hit table. Memory is $O(\text{problems})$,
not $O(\text{corpus})$; the audit run reported in
Section~\ref{sec:contamination-audit} held about $6$GB resident
and streamed the $12.9$\,B RedPajama-arXiv $13$-grams with $16$ workers on
one workstation.

\textbf{Slice selection and duplicate-source handling.}
The audit slice is RedPajama-arXiv. We initially included Dolma~v1.7's
\texttt{arxiv} subset, then verified that AI2 hosts it under URLs of the
form \texttt{olmo-data.org/dolma-v1\_7/redpajama-arxiv/arxiv-NNNN.json.gz}
that match the Together RedPajama manifest one-to-one---Dolma-arXiv adds
zero independent signal. We then included ProofPile-2's \texttt{arxiv}
subset and ran a dual-slice audit that returned an exact equality
$\text{hits}_{\mathrm{RP}}{=}\text{hits}_{\mathrm{PP}}$ for every problem
(zero asymmetries; matched-$n$-gram lists set-equal in $174/174$ problems
with hits). A document-level cross-reference confirmed the suspicion:
ProofPile-2-arXiv's \texttt{arXiv\_NNN.jsonl.zst} files are
RedPajama-arXiv's UUID-named \texttt{.jsonl.gz} files repackaged with the
same document order, the same per-document \texttt{arxiv\_id}, and the
same length. The Llemma technical report~\citep{azerbayev2024llemma}
states this in its corpus description. We therefore report the
release-defining labels against the single public arXiv slice for which the
paper establishes the audit boundary. Adding Pile-arXiv or a fresh
CommonCrawl-derived arXiv slice is left to future work.

\textbf{Cleaned-subset thresholds.}
$S_0$ is defined by $\text{hits}_{\mathrm{RP}}{=}0$ and $S_1$ by
$\text{docs}_{\mathrm{RP}}{\le}2$ distinct RedPajama-arXiv source documents.
The thresholds and the two floor rules
($\lvert S_0\rvert{<}200\Rightarrow$ switch headline to $S_1$;
$\lvert S_1\rvert{<}150\Rightarrow$ abandon cleaned-subset evaluation)
were pre-registered before the audit run; both are reported as not
triggered, with $\lvert S_1\rvert{=}3{,}985$ (doc-unique) exceeding the abort floor by
$26.6\times$.

\subsection{Uses}

\paragraph{Has the dataset been used for any tasks already?}
Yes. (a) The reference WirelessMathLM family ($0.5$\,B--$7$\,B Qwen-base
models trained with GRPO and binary verification rewards;
Section~\ref{sec:exp_grpo}). (b) The contamination-aware re-evaluation of the
reported model set on the test split and the cleaned subsets
$\mathcal{S}_0$/$\mathcal{S}_1$ (Section~\ref{sec:exp_contam}, Section~\ref{sec:contamination-audit}).
(c) Transfer evaluation of WirelessMathLM-trained checkpoints on five
general-mathematics benchmarks and four general-purpose benchmarks
(Section~\ref{sec:exp_transfer}). (d) A size-matched general-math GRPO
control evaluated on the WMB-XL test split as a domain-control baseline
(Section~\ref{sec:exp_transfer}).

\paragraph{Is there a repository linking to papers or systems using the dataset?}
A ``Models trained on / Papers using'' page is maintained in the code
repository (\url{https://github.com/LiXin97/WirelessMathBench-XL}).

\paragraph{What other tasks could the dataset be used for?}
Constraint-violation detection as a discriminative task; symbolic /
neuro-symbolic equation reconstruction; out-of-distribution probing
(MATH/OlympiadBench-trained models scored on a constraint-grounded
domain); contamination-aware benchmark methodology research using the
released audit pipeline as a template for adjacent technical domains.

\paragraph{Is there anything that should be considered when using the dataset for these tasks?}
Four caveats matter for downstream use. (i) Locked scores and comparisons are
defined under one semantic judge; an early cross-judge probe ($3.4$--$9.3\%$
absolute worst case per model) characterises judge sensitivity. (ii)
$\mathcal{S}_0/\mathcal{S}_1$ membership is a prompt-surface lexical-overlap
label against RedPajama-arXiv; it is not a measurement of memorisation,
answer-surface exposure, paraphrase exposure, or proprietary-corpus exposure.
OpenWebMath is an off-domain corroboration run, not the source of the released
cleaned subsets. (iii) The release split is not source-paper-disjoint; use the
grouped split-construction path before making source-paper-disjoint
generalisation claims about trained checkpoints. (iv) Two construction models
(DeepSeek-R1 extraction; GPT-4o filtering and rubric scoring) also appear as
evaluated rows, leaving a residual selection and style bias toward
R1-extractable content that per-record revision provenance does not quantify.

\paragraph{Are there tasks for which the dataset should not be used?}
Not as a wireless-engineering deployment instrument
without further validation: the dataset captures \emph{problems
appearing in the literature}, not the operational distribution of
field-deployed wireless systems. It is also not a safety / fairness
benchmark; the four implicit-constraint classes
(Section~\ref{sec:taxonomy_classes}) are physical / mathematical, not
social.

\subsection{Distribution}

\paragraph{Will the dataset be distributed to third parties?}
Yes, publicly on the Hugging Face Dataset Hub, subject to the
source-license restrictions below. The audit pipeline (reverse-probe $n$-gram
audit, $\mathcal{S}_0$/$\mathcal{S}_1$ construction scripts, the
materialized paper-disjoint sensitivity view, with identifiers for all $34$
items and text for the $13$ redistributable ones, the exact-format
lower-bound CSV, the paired-bootstrap evaluator, and per-model result
tables) is released alongside; training hyperparameters are documented in
Appendix~\ref{app:training_details}. Reference
WirelessMathLM checkpoints are released as separate model repositories on the
Hugging Face Hub.

\paragraph{How will the dataset be distributed?}
Through the Hugging Face Dataset Hub~\citep{huggingfacehub2024} (\url{https://huggingface.co/datasets/XINLI1997/WirelessMATHBench-XL})
for the dataset, GitHub (\url{https://github.com/LiXin97/WirelessMathBench-XL}) for code, and Hugging Face model repositories
for WirelessMathLM checkpoints. A Croissant~\citep{akhtar2024croissant}
metadata record is distributed with the dataset to enable ML-ready
discovery.

\paragraph{When will it be distributed?}
The v1.1 release is public with the camera-ready paper and supersedes
v1.0. Patch v1.1.1 adds per-item verdicts and result tables and changes no record.

\paragraph{Will the dataset be distributed under a license?}
Yes, per file, following the license of each record's source paper. We
harvested the source-paper license for all $836$ retained papers from
arXiv OAI-PMH metadata and release it as a per-paper manifest
(\texttt{source\_paper\_licenses.csv}) and as per-record
\texttt{source\_license} and \texttt{record\_license} fields. A record is
distributed only when its source license clearly permits redistribution of
derived text: $1{,}334$ records from CC~BY or CC0 sources are released under
CC~BY~4.0 (\texttt{problems\_cc\_by.jsonl}), $43$ records from CC~BY-SA sources
under CC~BY-SA~4.0 (\texttt{problems\_cc\_by\_sa.jsonl}), and $165$ records from
CC~BY-NC-SA sources under CC~BY-NC-SA~4.0
(\texttt{problems\_cc\_by\_nc\_sa.jsonl}). The remaining $2{,}485$ records,
derived from papers under the arXiv default non-exclusive distribution license
($2{,}160$), CC~BY-NC-ND ($323$), or no recorded license ($2$), are withheld and
listed by identifier only (problem, paper, and arXiv identifiers, split, format,
and license class) in \texttt{excluded\_records\_manifest.csv}. Withheld records
cannot be regenerated because construction used stochastic LLM calls. Of the
$800$ test problems, $310$ are distributed; this public test subset is the
externally reproducible evaluation view (Appendix~\ref{app:public_subset}).
Identifier-level audit metadata (contamination flags, split assignments,
and $\mathcal{S}_0$ membership by $n$ for all $4{,}027$ problems, and
evaluation result tables) contains no source-derived prose and is released under
CC~BY~4.0. Code is released under the MIT License. Source-paper text is not
redistributed.

\paragraph{Have any third parties imposed IP-based or other restrictions?}
The arXiv source papers carry the licenses chosen by their respective
authors (heterogeneous across the source pool). We do not redistribute
source-paper PDFs, LaTeX sources, or derivation paragraphs. The released
records contain derived mathematical problem statements, constraints,
ground-truth answers, source identifiers, and audit metadata under the
licenses above; correction, attribution, or takedown requests are
handled through the code repository issue tracker.

\paragraph{Do any export controls or other regulatory restrictions apply?}
None to our knowledge. The dataset is mathematical and contains no
deployment configurations, spectrum-access procedures, jamming tactics,
exploit instructions, cryptographic implementations, weapons-systems
specifications, or controlled-technology design files. Some wireless
communications research can have dual-use applications; this release is
limited to derived benchmark problems and audit metadata rather than
operational procedures.

\subsection{Maintenance}

\paragraph{Who will maintain the dataset?}
The paper authors; the primary maintainer is Xin Li.

\paragraph{How can the maintainer be contacted?}
Through the issue tracker at \url{https://github.com/LiXin97/WirelessMathBench-XL} or at \texttt{xin019@e.ntu.edu.sg}.

\paragraph{Is there an erratum?}
A \texttt{CHANGELOG.md} is maintained in the repository; errata are
posted there.

\paragraph{Will the dataset be updated?}
Yes, on a versioned schedule following semantic versioning
(MAJOR.MINOR.PATCH): new annotation rounds as MINOR bumps and
corrections as PATCH bumps. The contamination audit will be re-run at
each MAJOR version against any newly-accessible public pretraining
slices. The maintainers commit to a minimum maintenance horizon of
three years from the initial release (October 2025), covering issue triage,
audit re-runs against newly-released public corpora, and at least one
MINOR version bump per year.

Release v$1.1$ already ships a seed-$0$ paper-clustered split assignment and
a narrow answer-surface check at $n=13$. The planned v$1.2$ maintenance cycle
prioritizes a full answer-surface lexical audit
($\mathcal{S}_{0,\mathrm{target}}$), a stratified $200$-item three-rater
adjudication study for verifier calibration, and per-instance
constraint-taxonomy labels. These are roadmap items rather than claims of this
release.

\paragraph{If the dataset relates to people, are there applicable retention limits?}
Not applicable---no individual data subjects.

\paragraph{Will older versions be supported?}
Yes. Releases from v1.1 onward are tagged on the Hugging Face Hub, and
superseded versions are marked as such in the README. v1.0 has been
withdrawn because it included records whose source licenses do not permit
redistribution.

\paragraph{If others want to extend / augment the dataset, is there a mechanism?}
Yes. Pull requests against the repository will be reviewed by the
maintainers; a \texttt{CONTRIBUTING.md} (data-format spec, expected QA
process, audit-recompute requirement) is in the repository.

\section{Implicit-Constraint Taxonomy and Difficulty Structure}
\label{app:taxonomy}

Across the retained source papers, many wireless-math errors can be
described by four recurring implicit-constraint families. The families
are a qualitative taxonomy for interpreting problems and future audits;
the release does not include per-instance labels, deterministic class-specific
verifier paths, or prevalence estimates.

\subsection{Four Implicit-Constraint Classes}
\label{app:taxonomy_classes}

We organise the taxonomy by the underlying physical invariant being
violated, orthogonal to the subfield axis of Section~\ref{sec:stats}
(deep-learning and convex-optimisation papers both produce
power-non-negativity violations in different surface forms).

\noindent\textbf{Class 1 -- Power non-negativity.}
Powers, eigenvalues of covariance matrices, and rates are non-negative;
KKT-derived closed forms routinely admit negative roots and require a
water-filling projection $(\cdot)^{+}$ the problem text never states.

\noindent\textbf{Class 2 -- Complex-matrix dimensionality.}
$\mathbf{H} \in \mathbb{C}^{N \times M}$ carries implicit shape
constraints under Hermitian transpose, Kronecker product, and
inversion; silent $\mathbf{H}^{H}\mathbf{H}\!\leftrightarrow\!\mathbf{H}\mathbf{H}^{H}$
swaps break the next step's eigenvalue argument.

\noindent\textbf{Class 3 -- Log-domain validity.}
$\log(1+\gamma)$ assumes the SNR-like $\gamma$ stays in the domain
under the stated channel model; aggressive interference cancellation
can yield $1+\gamma \leq 0$ for which the achievability assumption
must be re-stated rather than the log evaluated.

\noindent\textbf{Class 4 -- Signal causality.}
Filters and precoders may use only past and present channel
realisations; conditioning on $\mathbf{h}[t+1]$ inside a $t$-time
decision is a structural violation that the algebra does not catch.

We use the four classes as qualitative construct labels in the
release. Corpus-wide prevalence estimates require a separate annotated
sample and are intentionally not reported here; the release keeps the
taxonomy definition separate from any future prevalence table.

\subsection{Difficulty Structure Across Task Formats}
\label{app:taxonomy_formats}

The three formats are not three difficulty levels of one problem; each
is a different probe of the constraint stack. \emph{MCQ} measures
detection of class-1 to class-4 violations given the equation's
surface form (distractors are designed to target constraint violations,
Section~\ref{sec:extraction}). \emph{Fill-in} graduates the amount of
constraint-consistent context at $25$\,/\,$50$\,/\,$75\%$ masking.
\emph{FEC} gives only variable
definitions, assumptions, and boundary conditions, and is therefore
the strictest test of constraint-driven reconstruction, while still
allowing formula-recall or surface-pattern shortcuts on some items. Per-class$\,\times\,$per-format gaps are a planned v$1.2$ audit
once instance-level class labels are added.

\section{Audit Corroboration and High-Overlap Analysis}
\label{app:audit-iterations}

\subsection{Why high-overlap problems do not equal memorised problems}
\label{app:audit:high-overlap-notation}

The contamination distribution has a long right tail: the highest-overlap
problem registers $159$ occurrence-count 13-gram hits across $75$ documents
against the RedPajama-arXiv slice, about $50\times$ the median of the
hit-bearing subset ($3$). It is a microservice latency-prediction problem, and
its retained matches are the graph declaration
$\mathcal{G}=(\mathcal{V},\mathcal{E})$ with its set definitions. The largest
flag in the test split (problem 5688, $49$ hits across $10$ documents) matches
the standard diagonal RIS reflection-matrix declaration. Both source papers
postdate the RedPajama-arXiv snapshot, so these hits cannot come from the papers
themselves; such patterns recur across preprints on many topics without implying
that a particular derivation is in pretraining. The methodological consequence is direct:
$13$-gram hit count is a narrow proxy for one lexical audit channel.
High hit count is not sufficient evidence of memorisation, and zero
hits do not rule out paraphrase, answer-only, alternate-version, or
closed-corpus exposure. A hit-weighted accuracy correction would
over-correct on notation-heavy subfields. The
paired-bootstrap headline of Section~\ref{sec:exp_contam} is
therefore computed on $\mathcal{S}_0$ (strict, zero hits) rather
than on a hit-weighted average of the full split.

\subsection{OpenWebMath off-domain corroboration run}
\label{app:audit:openwebmath}

We additionally ran the same reverse-probe audit against
OpenWebMath~\citep{paster2023openwebmath}, an independently-built
math-web pretraining corpus ($6.3$M documents, $7.85$\,B $13$-grams,
the same parallel implementation). Across the
$4{,}027$-problem corpus, $16$ problems ($0.40\%$) register any
overlap. OpenWebMath-only thresholds would yield
$|\mathcal{S}_0^{\mathrm{OWM}}| = 4{,}011$ and
$|\mathcal{S}_1^{\mathrm{OWM}}| = 4{,}021$. Of the $3{,}853$ problems in the
headline RP-arXiv $\mathcal{S}_0$, $3{,}848$ ($99.87\%$) also have zero detected hits against
OpenWebMath.

The OWM long-tail is dominated by canonical-formula recurrence rather
than problem-specific derivation text. Manual spot checks of the highest-hit
items find textbook expressions such as great-circle distance, discrete
Fourier-transform notation, the Hamilton quaternion identity, k-space
wave-vector notation, i.i.d. complex-Gaussian assumptions, and the
standard-normal inverse CDF. These are canonical mathematical surfaces that
recur across web sources, structurally analogous to the graph-declaration
boilerplate discussed in Appendix~\ref{app:audit:high-overlap-notation}. The inspected
matches are not problem-specific derivation steps or near-duplicates of WMB-XL
problem statements.

\subsection{Off-domain negative-control analysis (algebraic-stack)}
\label{app:audit:algebraic-stack}

To check that the reverse-probe pipeline is not pathologically
false-positive on unrelated mathematical text, we ran the same
audit against the ProofPile-2~\citep{azerbayev2024llemma}
\texttt{algebraic-stack} subset---an independently-built corpus of
mathematical \emph{source code} (Lean, Coq, Isabelle, Python,
Julia, Fortran), not arXiv text. The slice contains $3{,}440{,}226$
documents and $\sim$$3.02$B $13$-grams across $117$ files. The
audit completed in $412.2$\,s with the same parallel implementation
used for the RedPajama-arXiv run.

Across the $4{,}027$-problem corpus, $2$ problems ($0.05\%$) register any
overlap; all $4{,}025$ remaining problems return zero. Spot-checking traces the
matches to generic numerical-linear-algebra and algebraic identities that recur
across both wireless preliminaries and computational notebooks in the slice.
The overlap is incidental notation rather than evidence of the wireless problem
appearing in the algebraic-stack corpus. The near-zero rate---two orders of
magnitude below the RedPajama-arXiv hit fraction reported in
Section~\ref{sec:contamination-audit}---supports the interpretation that the
reverse-probe pipeline is not pathologically flagging unrelated mathematical
source-code text. It does not by itself calibrate false positives on unrelated
natural-language math.

\subsection{External reruns: portability and channel sensitivity}
\label{app:audit:external-reruns}

We reran the same fixed $13$-gram reverse-probe protocol on two external
benchmarks. Each run streamed $12.9$\,B RedPajama-ArXiv $13$-grams with
$16$ workers on one workstation.
On the $587$-item WirelessMathBench predecessor, the audit returns zero
detected hits. Its prompts expose about $29$ probe $13$-grams per item,
compared with about $127$ for WirelessMathBench-XL. This does not establish
absence of exposure; it shows that a short-prompt benchmark gives the lexical
channel fewer opportunities to fire.

On MATH-500~\citep{hendrycks2021measuring}, the audit flags $5/500$ items.
The released profile includes matched strings and source-document counts so
each flag can be inspected. The largest flag has $78$ matches across $12$
documents and includes a generic arithmetic sequence, illustrating why
evidence should accompany the binary label.

These runs demonstrate portability and inspectable output. They do not
validate the fixed threshold across domains or expand the channel to
paraphrase, target-answer, proprietary-corpus, or post-training exposure.

\section{Evaluation Robustness and Potential Bias}
\label{app:evaluation_robustness_app}

This appendix reports how the evaluation depends on its semantic judge. Every
locked verdict in Section~\ref{sec:exp} comes from one GPT-4.1-mini judge, which
receives the problem, the reference answer, and the model answer. Two separate
analyses characterise this dependence: an early cross-judge probe and a
rule-based exact-format check.

\paragraph{Early Cross-Judge Probe.}
The probe uses a two-stage grader on three high-performing response distributions.
\begin{itemize}
    \item \textbf{Stage 1 (Deterministic).} MCQ answers are graded by direct letter extraction. Fill-in / FEC answers are normalized by removing whitespace and LaTeX styling tokens (e.g., \texttt{\textbackslash mathbf}, \texttt{\textbackslash boldsymbol}) and then checked via exact string match.
    \item \textbf{Stage 2 (LLM Semantic Judge).} Open-form answers not settled by Stage 1 go to an LLM judge.
\end{itemize}
Stage 2 receives \textbf{GPT-5 base: 270/800 (33.75\%)}, \textbf{Gemini-2.5 Pro: 183/800 (22.88\%)}, and \textbf{Claude-Sonnet-4: 293/800 (36.63\%)} (Table~\ref{tab:evaluator_bias}). GPT-5 base is retained here only as a probe target; it is not a current leaderboard row.

\begin{table}[h]
\centering
\caption{Cross-evaluator agreement on the Stage-2 subset of the early probe.
GPT-5 (base) is a probe target, not a current leaderboard row.}
\label{tab:evaluator_bias}
\scriptsize
\resizebox{.85\linewidth}{!}{
\begin{tabular}{lcccc}
\toprule
\textbf{Model Evaluated} & \textbf{LLM Cases} & \textbf{GPT-4.1-mini} & \textbf{Claude-3-Haiku} & \textbf{Gemini-2.0-Flash} \\
\midrule
GPT-5 base & 270 & 41.11\% & 62.22\% & 42.22\% \\
Gemini-2.5 Pro & 183 & 53.01\% & 67.76\% & 56.28\% \\
Claude-Sonnet-4 & 293 & 41.64\% & 66.89\% & 45.05\% \\
\bottomrule
\end{tabular}
}
\end{table}

\paragraph{Cross-Evaluator Quantification.}
We re-graded the Stage-2 subset for these three probe targets using three independent judges (GPT-4.1-mini, Claude-3-Haiku, Gemini-2.0-Flash). Agreement rates on the Stage-2 subset differ by \textbf{up to 25.3 percentage points} across judges (Table~\ref{tab:evaluator_bias}); a judge reply that fails structured parsing is scored from the score written in the reply. Let $\rho$ denote the Stage-2 fraction and $\delta$ the maximum judge disagreement on that subset; then a conservative upper bound on overall-score variation is $\rho \cdot \delta$. This yields worst-case absolute impacts of:
\[
\begin{aligned}
\text{Gemini-2.5 Pro: } & 0.2288 \times 0.148 \approx 3.4\% \\
\text{GPT-5 base: } & 0.3375 \times 0.211 \approx 7.1\% \\
\text{Claude-Sonnet-4: } & 0.3663 \times 0.253 \approx 9.3\%.
\end{aligned}
\]
Cross-judge variation on this probe is \textbf{approximately 3.4--9.3\% absolute}, depending on the model. The range describes the two-stage probe rather than the locked tables, whose verdicts all come from one judge. It measures judge sensitivity, not ground-truth false negatives; a human/CAS adjudication study would measure verifier precision and recall directly.

\paragraph{Key Findings.} (1) \textbf{Judge dependence is explicit.} Locked scores and paired intervals are defined under one published judge, and this probe and the exact-format check characterise that dependence. (2) \textbf{The probe range is not verifier accuracy.} The 3.4--9.3\% absolute variation is a cross-judge sensitivity measure, not a human-ground-truth precision/recall estimate. (3) \textbf{Rankings are stable within this sensitivity analysis.} Across all three judges, relative ordering among the inspected top models remains unchanged; this supports, but does not prove, evaluator robustness for the full model set.

\paragraph{Exact-Format Lower-Bound Check.}
To read the cleaned-subset stability result without the semantic judge, we
also score archived response traces with an exact-format extractor only: MCQ letter
matching and exact comparison of extracted \texttt{\textbackslash boxed}
answers after lightweight \LaTeX{} normalisation. This check is stricter
than the semantic judge, so its absolute accuracies are lower-bound
format-match scores rather than replacement leaderboard numbers. Across the
five $16$k frontier rows and the three locked $2$k $7$B-scale reference rows,
the exact-format full-vs-$\mathcal{S}_0$ deltas lie in $[-0.21,+0.49]$ pp;
the CSV is released with the artifact as
\texttt{data/exact\_format\_lower\_bound.csv}.

\paragraph{Decoding-budget sensitivity.} The five $16$k rows of Tab.~\ref{tab:frontier_calibration}, including the GPT-$4.1$-mini judge-calibration row, are reported at the appropriate $16{,}384$-token budget; the remaining main-body rows are shown at the locked $2{,}048$-token protocol of Section~\ref{sec:exp_setup}. Two empirical observations motivate the dual-budget split: (i) GPT-$5$-mini gains $+3.00$\,pp on the same $800$ test ids when re-evaluated at $16$k vs $2$k, isolating the saturation-cluster gap as budget-limited rather than capability-limited; (ii) Gemini-$2.5$-Pro scores $41.25\%$ at the locked budget and $86.88\%$ at $16$k; at $2$k its visible responses average $435$ characters (Gemini-$2.5$-Flash: $2{,}261$) and $21/800$ are empty, so the reasoning budget consumes most of \texttt{max\_tokens}$=2048$ before the answer is written. Within each row, full and $\mathcal{S}_0$ are evaluated under the same token budget, so $\Delta(\text{full}-\mathcal{S}_0)$ is not a cross-budget comparison. We have not exhaustively rerun every row at $16$k, so cross-budget rank order is calibration-only. Truncation rates for the $16$k rows (\texttt{data/truncation\_16k.csv}) and 2k/16k deltas (\texttt{data/per\_model\_results.csv}) are released with the artifact.

\section{Detailed Evaluation Tables, Pairwise Bootstrap, and Inference Platform}
\label{app:cleaned_detail}

\begin{figure}[t]
\centering
\includegraphics[width=0.72\linewidth]{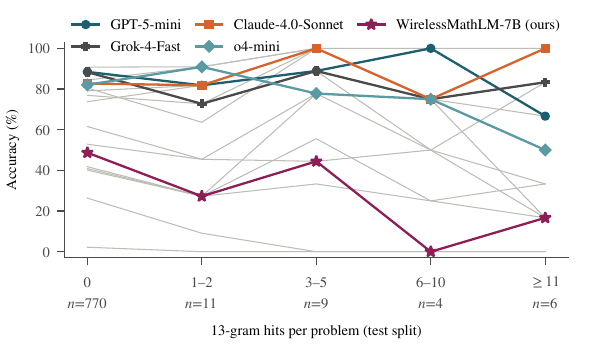}
\caption{Accuracy per overlap bucket for each reported row; the four
nonzero-hit buckets together hold $30$ test items.}
\label{fig:bucket_curves}
\end{figure}

This appendix provides the supporting evaluation material for Section~\ref{sec:exp}:
the locked $2$k per-format table, the per-overlap-bucket curves (Fig.~\ref{fig:bucket_curves}), the
WirelessMathLM-$7$B per-pair bootstrap row, and the inference-platform note. The per-format table uses the locked $18$-model
$2$k evaluation table; it is not a per-format expansion of the
appropriate-budget rows in Tab.~\ref{tab:frontier_calibration}.

\begin{table*}[t]
\centering
\caption{Locked $2$k per-format evaluation on the $800$-item release test split.
Rows are ranked by overall accuracy; bold marks the best score in each format.
$^{\dagger}$Budget-limited at $2$k (Appendix~\ref{app:evaluation_robustness_app}).}
\label{tab:per_format_leaderboard}
\scriptsize
\begin{tabular}{rlccccc}
\toprule
\# & \textbf{Model} & \textbf{Size} & \textbf{MCQ} & \textbf{Fill-in} & \textbf{FEC} & \textbf{Overall} \\
\midrule
 1 & GPT-$5.2$-chat                & --   & \textbf{93.98} & \textbf{89.08} & \textbf{92.67} & \textbf{90.75} \\
 2 & GPT-$5$-mini                  & --   & 92.48 & 86.97 & 88.48 & 88.25 \\
 3 & Grok-$4$-Fast                 & --   & 87.22 & 88.24 & 87.96 & 88.00 \\
 4 & DeepSeek-V$3.1$               & 671B & 85.71 & 85.71 & 82.20 & 84.88 \\
 5 & Claude-Sonnet-$4$             & --   & 84.21 & 84.03 & 79.58 & 83.00 \\
 6 & o$4$-mini                     & --   & 83.46 & 83.82 & 75.92 & 81.88 \\
 7 & Gemini-$2.5$-Flash            & --   & 81.20 & 81.09 & 78.53 & 80.50 \\
 8 & DeepSeek-R$1$                 & 671B & 81.20 & 80.88 & 73.30 & 79.12 \\
 9 & GPT-$4$o                      & --   & 79.70 & 77.73 & 73.30 & 77.00 \\
10 & Qwen$2.5$-$72$B-Instruct      & 72B  & 70.68 & 75.42 & 71.20 & 73.62 \\
\midrule
11 & Llama-$3.3$-$70$B-Instruct    & 70B  & 71.43 & 57.77 & 63.87 & 61.50 \\
12 & GPT-$5$-nano                  & --   & 66.92 & 54.83 & 36.65 & 52.50 \\
13 & WirelessMathLM-$7$B + GRPO (ours) & 7B & 48.12 & 50.63 & 40.84 & 47.88 \\
14 & Gemini-$2.5$-Pro$^{\dagger}$  & --   & 41.35 & 44.54 & 32.98 & 41.25 \\
15 & DeepSeekMath-$7$B-RL          & 7B   & 52.63 & 41.39 & 31.94 & 41.00 \\
16 & Qwen$2.5$-Math-$7$B-Instruct  & 7B   & 47.37 & 38.03 & 39.79 & 40.00 \\
17 & WirelessMathLM-$3$B + GRPO (ours) & 3B & 29.32 & 28.15 & 16.23 & 25.50 \\
18 & WirelessMathLM-$0.5$B + GRPO (ours) & 0.5B & 3.76 & 2.52 & 0.00 & 2.12 \\
\bottomrule
\end{tabular}
\end{table*}

\paragraph{Paired-difference protocol and WirelessMathLM-$7$B row.}
For every model pair $(i, j)$ in the locked $18$-model $2$k evaluation table we compute
$d_{ij}^{(\text{full})} = \text{acc}_i^{\text{full}} -
\text{acc}_j^{\text{full}}$ and $d_{ij}^{(\mathcal{S}_0)} =
\text{acc}_i^{\mathcal{S}_0} - \text{acc}_j^{\mathcal{S}_0}$, paired
by problem id within each subset, with $10{,}000$ bootstrap resamples
drawn at the problem level. This table excludes the
GPT-$4.1$-mini judge-calibration row and the selected $16$k reruns in
Tab.~\ref{tab:frontier_calibration}. The full calculation uses all $800$ test
items; the $\mathcal{S}_0$ calculation uses the $770$ zero-hit test
items. The full $\binom{18}{2}=153$-pair matrix is released as
\texttt{data/paired\_bootstrap\_matrix.csv}; the WirelessMathLM-$7$B row is in
Tab.~\ref{tab:wml7b_pairs}. Across the $153$ pairs, no point-estimate
pair ordering changes sign between the full and $\mathcal{S}_0$ splits.

\begin{table}[h]
\centering
\caption{Selected paired-bootstrap comparisons against WirelessMathLM-$7$B on
the strict zero-hit subset. The comparison uses $10{,}000$ resamples over the
$770$ paired $\mathcal{S}_0$ test items under the locked $2$k protocol.}
\label{tab:wml7b_pairs}
\scriptsize
\begin{tabular}{lcc}
\toprule
\textbf{Model $i$} & \textbf{$\Delta_{\mathcal{S}_0}$\,(pp)} & \textbf{$95\%$ CI\,(pp)} \\
\midrule
GPT-$5.2$-chat        & $+42.08$ & $[+38.31, +45.84]$ \\
GPT-$5$-mini          & $+39.74$ & $[+35.97, +43.51]$ \\
Grok-$4$-Fast         & $+39.61$ & $[+35.71, +43.51]$ \\
DeepSeek-V$3.1$       & $+35.71$ & $[+31.56, +39.87]$ \\
Claude-$4.0$-Sonnet   & $+34.03$ & $[+30.00, +37.92]$ \\
o$4$-mini             & $+33.38$ & $[+29.48, +37.40]$ \\
Gemini-$2.5$-Flash    & $+31.82$ & $[+27.79, +35.84]$ \\
DeepSeek-R$1$         & $+30.39$ & $[+26.36, +34.42]$ \\
GPT-$4$o              & $+28.18$ & $[+24.16, +32.21]$ \\
Qwen$2.5$-$72$B-Inst.   & $+25.06$ & $[+21.04, +28.96]$ \\
Llama-$3.3$-$70$B-Inst. & $+12.86$ & $[+8.57, +17.14]$ \\
GPT-$5$-nano          & $+\phantom{0}4.16$ & $[-0.13, +8.31]$ \\
\bottomrule
\end{tabular}
\end{table}

\paragraph{Inference-platform note.}
Closed-source frontier models are queried via vendor APIs (OpenAI,
Anthropic, Google, xAI) routed through OpenRouter under a consistent
hosted-inference request stack. Qwen$2.5$-$72$B-Instruct and
Llama-$3.3$-$70$B-Instruct are served via OpenRouter for reproducible hosted
inference under the same request stack. The WirelessMathLM family,
Qwen$2.5$-Math-$7$B-Instruct, DeepSeekMath-$7$B-RL, and Qwen$3$-$4$B
run locally on $4 \times \text{NVIDIA RTX 4090 D}$ ($48$\,GB each) via
\texttt{vllm 0.8.5} OpenAI-compatible server. Base-model GRPO
checkpoints (WirelessMathLM family, DeepSeekMath-$7$B-RL) are queried
via the raw-prompt \texttt{/v1/completions} endpoint to match
training-time prompting; chat-tuned baselines use the standard chat
endpoint. This load-bearing protocol choice affects every WirelessMathLM
and DeepSeekMath-RL row of Tab.~\ref{tab:main_results},
Tab.~\ref{tab:per_format_leaderboard}, and the per-pair CSV.

\section{Public Test Subset}
\label{app:public_subset}

Following the source-license-aware release (Appendix~\ref{app:datasheet},
Distribution section), $310$ of the $800$ test problems are distributed:
$273$ from CC~BY or CC0 sources, $9$ from CC~BY-SA sources, and $28$ from
CC~BY-NC-SA sources. Results on the full test split use the authors' complete
copy and cannot be reproduced externally, because the $490$ withheld test
records cannot be regenerated. Table~\ref{tab:public_subset} therefore
re-scores every evaluated row on the public subset from the same archived
per-item verdicts, without new model calls; CIs use $10{,}000$ bootstrap
resamples (seed $0$). Among the five $16$k frontier
rows, no pairwise ordering changes; among the $14$ locked-$2$k rows, one of
$91$ pairs changes order (DeepSeekMath-$7$B-RL and Qwen$2.5$-Math-$7$B-Instruct,
separated by $1.00$\,pp on the full split). The public subset is slightly
harder for most rows. It is the recommended view for externally comparable
results; the problem IDs are released as \texttt{public\_test\_ids.json}.

\begin{table}[!ht]
\centering
\caption{Accuracy on the full $800$-item test split and on the $310$-item
public test subset, re-scored from archived per-item verdicts. CI is a
$95\%$ bootstrap interval over public-subset items, $\Delta=\mathrm{Public}-\mathrm{Full}$,
D1--D14 are the locked-$2$k rows, and $\dagger$ marks construction-involved models.}
\label{tab:public_subset}
\scriptsize
\setlength{\tabcolsep}{3pt}
\begin{tabular}{llccccc}
\toprule
\textbf{ID} & \textbf{Model} & \textbf{Tok} & \textbf{Full \%} & \textbf{Public \%} & \textbf{$95\%$ CI} & \textbf{$\Delta$\,(pp)} \\
\midrule
-- & GPT-$5$-mini & $16$k & 91.25 & 90.32 & $[87.10, 93.55]$ & $-0.93$ \\
-- & GPT-$5.2$-chat & $16$k & 90.75 & 90.00 & $[86.45, 93.23]$ & $-0.75$ \\
-- & GPT-$4.1$-mini & $16$k & 89.62 & 89.35 & $[85.81, 92.58]$ & $-0.27$ \\
-- & Gemini-$2.5$-Pro & $16$k & 86.88 & 85.48 & $[81.61, 89.35]$ & $-1.39$ \\
-- & Gemini-$2.5$-Flash & $16$k & 86.50 & 84.19 & $[80.00, 88.06]$ & $-2.31$ \\
\midrule
D1 & Grok-$4$-Fast & $\phantom{0}2$k & 88.00 & 84.84 & $[80.65, 88.71]$ & $-3.16$ \\
D2 & DeepSeek-V$3.1$ & $\phantom{0}2$k & 84.88 & 83.87 & $[79.68, 87.74]$ & $-1.00$ \\
D3 & Claude-$4.0$-Sonnet & $\phantom{0}2$k & 83.00 & 82.26 & $[77.74, 86.45]$ & $-0.74$ \\
D4 & o$4$-mini & $\phantom{0}2$k & 81.88 & 79.35 & $[74.84, 83.55]$ & $-2.52$ \\
D5 & DeepSeek-R$1$$^{\dagger}$ & $\phantom{0}2$k & 79.12 & 77.74 & $[72.90, 82.26]$ & $-1.38$ \\
D6 & GPT-$4$o$^{\dagger}$ & $\phantom{0}2$k & 77.00 & 75.81 & $[70.97, 80.32]$ & $-1.19$ \\
D7 & Qwen$2.5$-$72$B-Instruct & $\phantom{0}2$k & 73.62 & 73.87 & $[69.03, 78.71]$ & $+0.25$ \\
D8 & Llama-$3.3$-$70$B-Instruct & $\phantom{0}2$k & 61.50 & 59.35 & $[53.87, 64.84]$ & $-2.15$ \\
D9 & GPT-$5$-nano & $\phantom{0}2$k & 52.50 & 51.29 & $[45.81, 56.77]$ & $-1.21$ \\
D10 & WirelessMathLM-$7$B & $\phantom{0}2$k & 47.88 & 46.45 & $[40.65, 51.94]$ & $-1.42$ \\
D11 & DeepSeekMath-$7$B-RL & $\phantom{0}2$k & 41.00 & 38.71 & $[33.23, 44.19]$ & $-2.29$ \\
D12 & Qwen$2.5$-Math-$7$B-Instruct & $\phantom{0}2$k & 40.00 & 40.65 & $[35.16, 46.13]$ & $+0.65$ \\
D13 & WirelessMathLM-$3$B & $\phantom{0}2$k & 25.50 & 25.16 & $[20.32, 30.00]$ & $-0.34$ \\
D14 & WirelessMathLM-$0.5$B & $\phantom{0}2$k & \phantom{0}2.12 & \phantom{0}2.26 & $[0.65, 4.19]$ & $+0.13$ \\
\bottomrule
\end{tabular}
\end{table}

\section{Source-Paper-Disjoint Sensitivity Analysis}
\label{app:paper-disjoint}

The release split is problem-level rather than
source-paper-disjoint. To make the residual dependence explicit, we form
the split-sensitivity slice
\[
\mathcal{S}_{\mathrm{paper\text{-}disjoint}}
= \{x \in \mathrm{test}: \mathrm{paper\_id}(x) \notin
\mathrm{paper\_id}(\mathrm{train})\}.
\]
This slice contains only $34$ test problems from $29$ source papers; the
remaining $766$ test problems come from papers that also contribute at
least one training problem. The slice is therefore underpowered and is
not a replacement leaderboard. Its purpose is to expose whether the
reported release-split behavior collapses on the small portion of the
test split that is source-paper-disjoint.

The supplementary release materializes this split-sensitivity view and includes a
seed-$0$ grouped split assignment with $3{,}222$ train records and $805$ test
records under zero source-paper overlap. These files make the release-split
dependence auditable, but they do not convert the problem-level release split
into a paper-disjoint training benchmark. New model claims on that view require
retraining and reevaluation under the rebuilt split.

Using the locked $18$-model $2$k evaluation table, per-model accuracy on the
$34$-item slice is noisy. This table is the split-sensitivity view used
for pairwise checks, not the appropriate-budget calibration view in
Tab.~\ref{tab:frontier_calibration}. Selected rows are shown in
Tab.~\ref{tab:paper_disjoint_sensitivity}; positive deltas on this $34$-item
slice should not be read as a replacement leaderboard.

This split-sensitivity view does not repair the release split. Any future claim about
paper-disjoint training generalisation should use a grouped split such as the
released seed-$0$ assignment, retrain the reference models, and rerun the
contamination-aware evaluation under that clustered partition.

\section{Preview-Model Exclusion Rationale}
\label{app:preview-instability}

Two Gemini-3 preview releases tested during frontier calibration returned
a non-trivial fraction of empty response bodies (\verb|candidate=""| with
\verb|finish_reason=""|) under our standard call configuration. We
document the rate here and exclude both rows from the audited calibration
tables (Section~\ref{sec:exp_contam}, Tab.~\ref{tab:main_results}) to avoid
contaminating the cross-model comparison with what is, on the available
evidence, an API-side artifact rather than a model-quality signal.

\paragraph{Diagnosis.}
For \texttt{gemini-3.1-pro-preview}, the full $n{=}800$ run returned an
empty candidate body on $196/800$ calls ($24.5\%$) with an empty
\texttt{finish\_reason} string. This differs from ordinary token-budget
exhaustion, where empty outputs co-occur with \texttt{finish\_reason=length}.
Conditional on a non-empty body, accuracy was $89.6\%$. The same failure mode
appeared for \texttt{gemini-3-flash-preview}, with $350/800$ empty bodies
($43.8\%$) and $89.3\%$ conditional accuracy. We report these conditional
accuracies only to justify exclusion; the missing-response rates make the raw
rows unsuitable for the main cross-model comparison.

\paragraph{Reported numbers.}
Both models are reported in Tab.~\ref{tab:preview-instability} and
excluded from the audited calibration tables. We do not use preview-model
conditional accuracy as evidence for any paper claim.

\begin{table}[h]
\centering
\caption{Gemini-3 preview rows excluded from the headline calibration tables.
Both preview APIs return many empty response bodies; conditional accuracy is
computed only over non-empty responses.}
\label{tab:preview-instability}
\scriptsize
\begin{tabular}{lrrrr}
\toprule
Model & $n_{\text{total}}$ & $n_{\text{valid}}$ & Empty-body \% & Acc. on $n_{\text{valid}}$ \\
\midrule
\texttt{gemini-3.1-pro-preview}    & 800 & 604 & $24.5\%$ & $89.6\%$ \\
\texttt{gemini-3-flash-preview}    & 800 & 450 & $43.8\%$ & $89.3\%$ \\
\bottomrule
\end{tabular}
\end{table}

\section{Per-Subdomain Performance Analysis}
\label{app:per_subdomain}

We analyze performance across 20 wireless subdomain tags in the
800-problem test set. Tags are multi-label, so category counts do not
sum to 800. This is a descriptive release-split breakdown, not a
paper-disjoint or memorisation analysis. Tags come from a GPT-4.1-mini,
$20$-category, multi-label classification of the test problems; the table
is the retained aggregate output, and per-problem tags were not saved.

\begin{table}[h]
\centering
\caption{Accuracy by multi-label subdomain tag on the $800$-problem test set.
Base and +GRPO are the Qwen$2.5$-$7$B rows of Tab.~\ref{tab:grpo_scale}
(training-time greedy evaluation).}
\label{tab:subdomain_performance}
\scriptsize
\resizebox{.95\linewidth}{!}{
\begin{tabular}{lcccccc}
\toprule
\textbf{Category} & \textbf{Count} & \textbf{\%} & \textbf{Base (7B)} & \textbf{+GRPO} & \textbf{$\Delta$ Abs} & \textbf{$\Delta$ Rel} \\
\midrule
Channel Estimation & 160 & 20.0\% & 22.50\% & 36.25\% & +13.75\% & +61.1\% \\
Beamforming & 113 & 14.1\% & 18.58\% & 43.36\% & +24.78\% & +133.3\% \\
Resource Allocation & 110 & 13.8\% & 21.82\% & 39.09\% & +17.27\% & +79.2\% \\
MIMO/Massive MIMO & 104 & 13.0\% & 23.08\% & 43.27\% & +20.19\% & +87.5\% \\
Information Theory & 103 & 12.9\% & 21.36\% & 46.60\% & +25.24\% & +118.2\% \\
RIS/IRS & 98 & 12.3\% & 17.35\% & 33.67\% & +16.33\% & +94.1\% \\
Machine Learning & 86 & 10.8\% & 25.58\% & 50.00\% & +24.42\% & +95.5\% \\
OFDM/OFDMA & 50 & 6.3\% & 18.00\% & 28.00\% & +10.00\% & +55.6\% \\
ISAC & 49 & 6.1\% & 24.49\% & 36.73\% & +12.24\% & +50.0\% \\
Convex Optimization & 49 & 6.1\% & 26.53\% & 36.73\% & +10.20\% & +38.5\% \\
Deep Learning & 44 & 5.5\% & 27.27\% & 43.18\% & +15.91\% & +58.3\% \\
Stochastic Geometry & 40 & 5.0\% & 35.00\% & 32.50\% & -2.50\% & -7.1\% \\
UAV Communications & 32 & 4.0\% & 12.50\% & 31.25\% & +18.75\% & +150.0\% \\
Edge Computing & 28 & 3.5\% & 7.14\% & 39.29\% & +32.14\% & +450.0\% \\
Reinforcement Learning & 28 & 3.5\% & 32.14\% & 35.71\% & +3.57\% & +11.1\% \\
Channel Coding & 20 & 2.5\% & 25.00\% & 45.00\% & +20.00\% & +80.0\% \\
Federated Learning & 19 & 2.4\% & 26.32\% & 36.84\% & +10.53\% & +40.0\% \\
NOMA & 19 & 2.4\% & 15.79\% & 26.32\% & +10.53\% & +66.7\% \\
Semantic Communications & 13 & 1.6\% & 15.38\% & 61.54\% & +46.15\% & +300.0\% \\
Cell-Free & 9 & 1.1\% & 11.11\% & 33.33\% & +22.22\% & +200.0\% \\
\midrule
\textbf{Overall (all $800$ problems)} & 800 & 100\% & 21.88\% & 39.50\% & \textbf{+17.62\%} & \textbf{+80.5\%} \\
\bottomrule
\end{tabular}
}
\end{table}

\textbf{Interpretation.} The subdomain rows are coverage summaries, not
class-level evidence. Among tags with at least $30$ test problems, GRPO
point estimates mostly improve, with Stochastic Geometry as the exception;
tags below this threshold (including Semantic Communications and Cell-Free)
are too small for relative-gain interpretation. These rows do not
distinguish transferable reasoning from same-source-paper dependence, tag
imbalance, verifier effects, or memorisation.

\section{Detailed Qualitative Analysis}
\label{app:qualitative_analysis}

We provide selected case studies illustrating correct behaviors observed
in WirelessMathLM-7B outputs. These examples are qualitative illustrations, not frequency
estimates or evidence that the behavior was caused by GRPO.

\subsection{Case Study 1: Domain-Specific Operations}
In one downlink MISO problem (Question ID 6759), the model separated the
received symbol at user $k$ into the desired signal, multi-user
interference, and channel noise, and derived the correct expression:
\begin{equation}
[\mathbf{y}_k]_n = \mathbf{h}_k^H \mathbf{v}_k [\mathbf{z}_k]_n + \sum_{m \neq k} \mathbf{h}_k^H \mathbf{v}_m [\mathbf{z}_m]_n + \sqrt{\sigma^2}\, n
\end{equation}
This response illustrates the Hermitian transpose of the channel vector
($\mathbf{h}_k^H$), an interference sum that excludes the intended user,
and noise scaling ($\sqrt{\sigma^2}$), which are distinct from general
algebraic rules.

\subsection{Case Study 2: Method Justification}
Some correct solutions include explicit rationales for chosen approaches.
In a hybrid-beamforming problem for a UE-mounted RIS system (Question ID
15115), the model first stated its approach, ``we need to analyze the
signal flow in the hybrid beamforming architecture of the UE-mounted RIS
system'', and then wrote the
received signal as
\begin{equation}
\mathbf{y} = \mathbf{W}_D^H \mathbf{W}_A^H \mathbf{H} \mathbf{F}_A \mathbf{F}_D \mathbf{x} + \mathbf{W}_D^H \mathbf{W}_A^H \mathbf{n},
\end{equation}
naming the role of each factor: digital and analog combiners, effective
channel, analog and digital precoders. The three distractors permute
these same factors, so the signal-flow ordering is what identifies the
correct option.

\subsection{Case Study 3: Physical Intuition}
In an edge-learning transmission-energy problem (Question ID 7768), the
masked terms are the energy coefficient and the two denominator factors
of $\mathbf{E}(b, t; h_k) = \lambda\, b^\ell / (t^{\ell-1} h_k)$. The
solution places the channel gain by physical reasoning: ``The channel
gain is in the denominator because it affects the energy consumption. A
higher channel gain (i.e., a stronger signal) typically reduces the
energy required for transmission.''

\Needspace{20\baselineskip}
\section{Representative Solution Examples from WirelessMathLM-7B}
\label{app:solution_examples}

This appendix presents detailed examples of solutions generated by WirelessMathLM-7B, organized by question type and quality level.

\subsection{High-Quality Solution Examples}

\subsubsection{Multiple Choice Question: Hybrid Beamforming Signal Model}
\textbf{Question ID:} 15115. \textbf{Correct Answer:} A. \textbf{Model Response:} the expression of option A.

\begin{tcolorbox}[colback=correctgreen!10,colframe=correctgreen,title=WirelessMathLM-7B Solution (Truncated),breakable]
To determine the correct expression that should replace [MASK] in the given equation, we need to analyze the signal flow in the hybrid beamforming architecture of the UE-mounted RIS system. The received signal vector $\mathbf{y}$ is given by:
\[
\mathbf{y} = \mathbf{W}_D^H \mathbf{W}_A^H \mathbf{H} \mathbf{F}_A \mathbf{F}_D \mathbf{x} + \mathbf{W}_D^H \mathbf{W}_A^H \mathbf{n}
\]
Here, $\mathbf{y}$ is the received symbol vector after combining, $\mathbf{x}$ is the transmitted signal vector, $\mathbf{n}$ is the additive white Gaussian noise vector, $\mathbf{W}_D^H$ is the digital combiner, $\mathbf{W}_A^H$ is the analog combiner, $\mathbf{H}$ is the total effective channel, $\mathbf{F}_A$ is the analog precoder, and $\mathbf{F}_D$ is the digital precoder. Thus, the expression that should replace [MASK] is:
\[
\boxed{\mathbf{W}_D^H \mathbf{W}_A^H \mathbf{H} \mathbf{F}_A \mathbf{F}_D \mathbf{x}}
\]
\end{tcolorbox}

\textbf{Analysis:} The solution states its approach, names the role of every factor, and keeps them in signal-flow order, which separates the correct option from the three distractors that permute the same factors.

\subsubsection{Fill-in-the-Blank (100\%): Downlink MISO Received Signal}
\textbf{Question ID:} 6759. \textbf{Correct Answer:} $[\mathbf{y}_k]_n = \mathbf{h}_k^H \mathbf{v}_k [\mathbf{z}_k]_n + \sum_{m=1, m\neq k}^{K} \mathbf{h}_k^H \mathbf{v}_m [\mathbf{z}_m]_n + \sigma n$.

\begin{tcolorbox}[colback=correctgreen!10,colframe=correctgreen,breakable,title=WirelessMathLM-7B Solution (Truncated)]
To derive the equation for the $n$-th received symbol at user $k$, we need to consider the contributions from the desired signal, multi-user interference, and channel noise. The received signal at user $k$ due to the desired signal is given by $\mathbf{h}_k^H \mathbf{v}_k [\mathbf{z}_k]_n$. For each user $m$ (where $m \neq k$), the interference is given by $\mathbf{h}_k^H \mathbf{v}_m [\mathbf{z}_m]_n$. The received signal at user $k$ due to the channel noise is $\sqrt{\sigma^2} n$. Thus, the final answer is:
\[
\boxed{[\mathbf{y}_k]_n = \mathbf{h}_k^H \mathbf{v}_k [\mathbf{z}_k]_n + \sum_{m \neq k} \mathbf{h}_k^H \mathbf{v}_m [\mathbf{z}_m]_n + \sqrt{\sigma^2} n}
\]
\end{tcolorbox}

\textbf{Analysis:} The solution separates desired signal, multi-user interference, and noise, applies the Hermitian transpose to the channel vector, and excludes the intended user from the interference sum.

\subsubsection{Fill-in-the-Blank (50\%): Wavelet Scaling and Translation}
\textbf{Question ID:} 11142. \textbf{Correct Answer:} $\frac{1}{\sqrt{a}}$ and $t - b$.\nopagebreak

\begin{tcolorbox}[colback=correctgreen!10,colframe=correctgreen,title=WirelessMathLM-7B Solution (Truncated)]
To ensure energy normalization in the context of the continuous wavelet transform, the wavelet function $\psi(t)$ must be scaled such that the integral of its square over all time is equal to 1. The energy normalization factor is $\frac{1}{\sqrt{a}}$, and the translated time argument is $t - b$. Therefore, the wavelet function $\psi_{a,b}(t)$ can be written as:\par\centerline{$\psi_{a,b}(t) = \frac{1}{\sqrt{a}} \cdot \psi\left( \frac{t - b}{a} \right)$}
Thus, the final answer is: $\boxed{\frac{1}{\sqrt{a}}, t - b}$
\end{tcolorbox}

\textbf{Analysis:} Correct identification of the energy-normalization factor and the translated time argument.

\subsection{Error Analysis Examples}

\subsubsection{Missing-Factor Error}
\textbf{Question ID:} 1631. \textbf{Type:} Missing Factor.
\textbf{Correct:} $2R_L$. \textbf{Model:} $R_L$.

\begin{tcolorbox}[colback=red!10,colframe=red,title=Error Example: Missing Factor (Truncated)]
The output power $P_{\mathrm{out}}$ delivered to the load resistor $R_L$ is given by the formula: $P_{\mathrm{out}} = \frac{V_{\mathrm{out}}^2}{R_L}$. Here, $V_{\mathrm{out}}$ is the peak voltage of the sinusoidal output signal, and $R_L$ is the load resistance. Therefore, the denominator of the output power expression is $\boxed{R_L}$.
\end{tcolorbox}

\textbf{Error Analysis:} Correct power--voltage--resistance relation but a missing peak-to-RMS factor.

\subsubsection{Conceptual Misunderstanding Error}
\textbf{Question ID:} 16144. \textbf{Type:} Conceptual Misunderstanding.\\
\textbf{Correct:} $\sum_{m=1}^{M} \sum_{k=1}^{K_i} \int_0^{T_{s_k}} \tau_{g_m}(t) P(t)\, dt$.

\begin{tcolorbox}[colback=red!10,colframe=red,title=Error Example: Conceptual Misunderstanding (Truncated)]
The total transmission energy $E_t$ can be expressed as the sum of the transmission energy for each GBS that the UAV is transmitting to. This can be written as:\par\centerline{$E_t = \sum_{i=1}^{M} \sum_{k=1}^{K_i} T_{s_k} \cdot \tau_{g_m}(t) \cdot P(t)$}
\end{tcolorbox}

\textbf{Error Analysis:} Discrete sums in place of the time integral, with inconsistent summation indices.

\subsubsection{MCQ Selection Error}
\textbf{Question ID:} 17370. \textbf{Type:} Multiple Choice Selection.
\textbf{Correct:} A. \textbf{Model:} $\sum_{n=1}^{j-1} \alpha_{n,k} |h_{i,k}|^2 + \rho$, after restating the options.\nopagebreak

\begin{tcolorbox}[colback=red!10,colframe=red,title=Error Example: MCQ Selection (Truncated)]
\begin{enumerate}\setcounter{enumi}{2}
\item \textbf{Noise Power}: This is the power of the noise at user $i$. It is given by $\rho$, which is the transmit signal-to-noise ratio (SNR) in dimensionless form.
\end{enumerate}
\end{tcolorbox}

\textbf{Error Analysis:} Correct interference structure paired with an inverted noise normalization.

\section{Training Implementation Details}
\label{app:training_details}

\subsection{Mathematical Formulation of GRPO}

We use Group Relative Policy Optimisation~\citep{shao2024deepseekmath}
with a binary verifiable reward. For a query $x \sim P(X)$, we sample a
group of $G$ responses $\{y_i\}_{i=1}^G$ from the previous policy
$\pi_{\theta_{\text{old}}}$ and assign each a scalar reward
$R_i = 0.1\,r_{\text{format}}(y_i) + 0.9\,r_{\text{accuracy}}(x, y_i)$,
where $r_{\text{accuracy}} \in \{0,1\}$ is the terminal verdict of the
hierarchical training verifier (deterministic match
$\to$ canonicalisation $\to$ bounded \textsc{GPT-4.1-mini} fallback)
and $r_{\text{format}} \in \{0,1\}$ checks tag well-formedness.
Group-standardised advantages
\begin{equation}
A_i \;=\; \frac{R_i - \mu_R}{\sigma_R},
\qquad
\mu_R = \tfrac{1}{G}\!\sum_{j=1}^{G} R_j,\quad
\sigma_R = \mathrm{std}(\{R_j\}_{j=1}^{G})
\end{equation}
remove the need for a learned value baseline. The training objective
adds an explicit KL regulariser against a frozen reference policy
$\pi_{\text{ref}}$ initialised from the base checkpoint:
\begin{equation}
\begin{aligned}
\mathcal{J}_{\text{GRPO}}(\theta) \;=\; \mathbb{E}_{x,\, \{y_i\}}
\Bigg[
\frac{1}{G}\sum_{i=1}^{G}
\min\!\Big(\, & r_i(\theta)\, A_i,\;
\mathrm{clip}\!\big(r_i(\theta),\, 1-\epsilon,\, 1+\epsilon\big)\, A_i \Big) \\
- \,\beta\,& D_{\mathrm{KL}}\!\left[\pi_{\theta}(\cdot \mid x)\,\big\|\,\pi_{\text{ref}}(\cdot \mid x)\right]
\Bigg],
\end{aligned}
\end{equation}
where $r_i(\theta) = \pi_\theta(y_i \mid x) / \pi_{\theta_{\text{old}}}(y_i \mid x)$,
$\epsilon = 0.2$, $\beta = 0.01$, and $G = 8$. The Qwen2.5 runs initialise from
the base checkpoints and the Qwen3-4B run from the released post-trained
Qwen3-4B checkpoint; no run uses an SFT warm-start. Training temperature is
$T = 1.0$ and evaluation $T = 0.6$.

\subsection{Hyperparameters and Infrastructure}
Table~\ref{tab:hyperparams} lists the detailed hyperparameters used for training WirelessMathLM. We utilized a cluster of 4 NVIDIA RTX 4090 D GPUs. Training took 28.4 hours for 3B and 53.4 hours for 7B.

\begin{table}[h]
\centering
\caption{Hyperparameters for GRPO Training}
\label{tab:hyperparams}
\scriptsize
\resizebox{.55\linewidth}{!}{
\begin{tabular}{lc}
\toprule
\textbf{Hyperparameter} & \textbf{Value} \\
\midrule
Base Model & Qwen2.5-Base \{0.5B, 3B, 7B\} \\
           & + Qwen3-4B$^{\dagger}$ \\
Optimizer & AdamW \\
Learning Rate & $1 \times 10^{-6}$ \\
Weight Decay & 0.1 \\
LR Scheduler & Cosine Annealing \\
Global Batch Size & 64 \\
Group Size ($G$) & 8 \\
KL Coefficient ($\beta$) & 0.01 \\
Clip Range ($\epsilon$) & 0.2 \\
Max Context Length & 2048 \\
Epochs & 40 \\
\bottomrule
\end{tabular}
}
\end{table}

\noindent$^{\dagger}$ Qwen3-4B is the released post-trained checkpoint, not
Qwen3-4B-Base; its training prompt appends \texttt{/no\_think}. It uses the
same hyperparameter set as Qwen2.5-7B above; no Qwen3-specific hyperparameter
sweep was performed.

\subsection{Training-Time Checks and SFT Ablation}

Training used EasyR1. The release has no separate validation split: during
training the test split was evaluated every $5$ steps, and the step budget
was extended once, from $120$ to $240$ steps, while test accuracy was still
rising. Every GRPO row of Table~\ref{tab:grpo_scale} reports the final
checkpoint (step $240$), and the SFT row reports epoch $6$;
all rows use the training-time evaluation, a protocol separate from the
locked $T{=}0.6$ evaluation of Section~\ref{sec:exp}. The Qwen$2.5$-$3$B row
evaluates the released step-$240$ checkpoint (WirelessMathLM-$3$B) with the
same evaluation script, generation settings, and GPT-4.1-mini judge (now
accessed through OpenRouter); evaluating the $7$B step-$240$ checkpoint the
same way gives $40.00\%$ against the tabulated $39.50\%$. GRPO raises overall
accuracy by $+13.25$ and $+17.62$ points at $3$B and $7$B. The $0.5$B checkpoint moves by $+1.49$ and stays near the floor, and a
single Qwen$3$-$4$B run improves by $+4.24$. The Qwen$2.5$-$3$B SFT ablation
reaches $19.13\%$ at epoch $6$ under the same protocol, below the
$25.62\%$ GRPO row; SFT and GRPO use different
feedback signals, so this is a contextual comparison rather than a
compute-normalised one.

\begin{table}[h]
\centering
\caption{Training-time accuracy (\%) before and after GRPO at the final
checkpoint, with an SFT ablation on Qwen$2.5$-$3$B.}
\label{tab:grpo_scale}
\small
\begin{tabular}{lccc}
\toprule
\textbf{Base} & \textbf{Base} & \textbf{+GRPO} & \textbf{$\Delta$ (rel.)} \\
\midrule
Qwen$2.5$-$0.5$B  & 13.38 & 14.87 & +1.49\,\,($+11\%$)  \\
Qwen$2.5$-$3$B    & 12.37 & 25.62 & +13.25 ($+107\%$)   \\
Qwen$3$-$4$B$^{\dagger}$ & 13.63 & 17.87 & +4.24\,\,($+31\%$)  \\
Qwen$2.5$-$7$B    & 21.88 & 39.50 & +17.62 ($+81\%$)    \\
\midrule
\multicolumn{4}{l}{\textit{SFT ablation, Qwen$2.5$-$3$B}} \\
\quad SFT (epoch 6) & --- & 19.13 & --- \\
\bottomrule
\end{tabular}
\end{table}

\section{Use Cases and Scope Boundaries}
\label{app:use_cases}

\noindent\textbf{Wireless engineering copilot.}
The three task formats map to three copilot scenarios with monotonically
increasing engineer-side trust requirements: MCQ to \emph{derivation
review} (flag a constraint-violating continuation among four), Fill-in to
\emph{specification-driven completion}, FEC to \emph{drafting from a
specification}. WirelessMathLM-$7$B reaches a $47.88\%$ release-split
point estimate; the two competing-family $7$B math-specialist baselines
(Qwen$2.5$-Math-$7$B-Instruct and DeepSeekMath-$7$B-RL) sit
$6.88$--$7.88$\,pp below it, with paired $95\%$ intervals under the
same judge of $[+2.88,+11.00]$ and $[+3.88,+11.88]$. The model still does not approach the
$2026$ frontier calibration rows, which cluster between $86.5\%$ and
$91.3\%$. The implication for
deployment is a layered one: useful as a
review aid where a human keeps the pen, useful as a draft generator on
specification-driven completion, not yet deployment-ready on full
drafting from a specification.

\noindent\textbf{Implicit-constraint reasoning beyond wireless.}
The four classes of Section~\ref{sec:taxonomy_classes} are a descriptive
feature of the dataset rather than a load-bearing contribution of this
paper, and they are not wireless-specific: power non-negativity
generalises to feasibility-region projection; complex matrix
dimensionality recurs in quantum-circuit and electromagnetics
derivations; log-domain validity in thermodynamic partition functions;
signal causality in non-anticipation constraints. We flag the
generalisation as an opening for follow-up work in neighbouring
technical-math domains, not as a measured cross-domain claim of the
present paper.

\noindent\textbf{Scope, threats, future work.}
Table~\ref{tab:threats_to_validity} consolidates the main validity
boundaries. The table is descriptive: it adds no audit channel beyond the
released public-corpus checks and no paper-disjoint training-generalisation
claim beyond the materialized sensitivity views.

\begin{table}[t]
\centering
\caption{Threats to validity and how they are handled in the release.}
\label{tab:threats_to_validity}
\scriptsize
\setlength{\tabcolsep}{3pt}
\begin{tabular}{>{\raggedright\arraybackslash}p{0.20\linewidth}>{\raggedright\arraybackslash}p{0.28\linewidth}>{\raggedright\arraybackslash}p{0.42\linewidth}}
\toprule
\textbf{Threat} & \textbf{Current treatment} & \textbf{Remaining limitation / next step} \\
\midrule
Prompt-surface exact overlap & Reverse-probe RedPajama-arXiv audit with fixed $13$-gram threshold; release ships $\mathcal{S}_0/\mathcal{S}_1$ labels and rebuild scripts. & The bound is channel-specific and threshold-specific; users can tighten or replace the released filter. \\
Answer-surface exposure & \texttt{correct\_answer} is deliberately excluded from the prompt-surface audit; a narrow answer-side $13$-gram lower-bound check is reported separately. & A fuller target-equation / answer-surface audit is needed before claiming anything about answer exposure. \\
Paraphrase or near-duplicate exposure & Limitations are stated in the main text and audit appendix. & Approximate near-duplicate, prompt-rewrite, or embedding-style probes are complementary future audit channels, not results of this release. \\
Closed corpora and alternate arXiv mirrors & RedPajama-arXiv is the release-defining public slice; OpenWebMath is off-domain corroboration and algebraic-stack is a negative control. & Closed/proprietary mixtures and unprobed mirrors can contain exposure that public-corpus labels miss. \\
Verifier / judge dependence & Every locked verdict comes from the GPT-4.1-mini judge; an early cross-judge probe and an exact-format lower-bound check are reported. & The $3.4$--$9.3\%$ probe range is not a bound on the locked tables or a human/CAS precision-recall estimate; held-out verifier calibration remains future work. \\
Release-split dependence & All records retain \texttt{paper\_id}; a $34$-item source-paper-disjoint sensitivity view is materialized in the release. & Training rows are release-split learnability checks, not source-paper-disjoint generalisation estimates. \\
\bottomrule
\end{tabular}
\end{table}

The practical reading is simple: use $\mathcal{S}_0/\mathcal{S}_1$ for the
fixed prompt-surface audit channel, the exact-format and cross-judge analyses
for verifier sensitivity, and the grouped split scripts for paper-disjoint
retraining. Broader exposure probes and held-out verifier calibration are
roadmap items, not results of this release.

\medskip

\noindent\textbf{The audit returns no frontier-rank-relevant signal under the fixed channel.}
\label{sec:exp_contam_unexpected}
The contamination audit and re-evaluation were instrumented to test whether
detected RedPajama-arXiv prompt-surface overlap changes the frontier-vs-reference
picture. It does not under the fixed channel: every audited full-vs-$\mathcal{S}_0$
row stays within $[-0.86,+0.46]$\,pp, bucket-level Spearman correlations remain
small ($|\rho|\le0.11$), and the released $\binom{18}{2}=153$ pair matrix has no
point-estimate sign change between the full and cleaned splits. We report this
as the practical bound returned by the fixed $13$-gram lexical audit against the
audited public corpora, not as a claim about memorisation, paraphrase exposure,
or closed pretraining mixtures. The released audit infrastructure is the
starting point for those stronger probes.

\section{Use of Large Language Models}
\label{app:llm_disclosure}

We disclose that LLMs were utilized as tools in various stages of this research project. The final responsibility for all content, including its accuracy and originality, rests with the human authors.

\begin{itemize}
    \item \textbf{Writing and Editing:} LLMs were used to assist with improving the grammar, clarity, and style of the manuscript. The authors reviewed and edited all LLM-generated text to ensure it accurately reflects our research and findings.

    \item \textbf{Literature Discovery:} LLMs were employed to help summarize related work and accelerate the literature discovery process, assisting in identifying relevant prior research in mathematical reasoning and domain adaptation.

    \item \textbf{Dataset Curation Pipeline:} As detailed in Appendix~\ref{app:llm_quality_assessment}, LLMs were integral to the construction of the \textbf{WirelessMathBench-XL} dataset. Specifically:
    \begin{itemize}
        \item \textbf{Paper Filtering:} GPT-4o was used to perform an initial filtering of candidate papers to identify those with substantial mathematical content relevant to wireless communications.
        \item \textbf{Content Extraction:} DeepSeek-R1 was used to extract structured mathematical models from the LaTeX source of selected papers.
        \item \textbf{Automated Quality Assessment:} GPT-4o was used as part of a multi-tier quality assurance framework to perform initial automated evaluations of generated questions.
    \end{itemize}

    \item \textbf{Evaluation:} For our evaluation metric, GPT-4.1-mini was used to grade each model response against its reference answer.
\end{itemize}

In all instances, LLM outputs were critically reviewed, validated, and verified by the authors. We take full responsibility for the claims, results, and conclusions presented in this paper.

\newpage
\section*{NeurIPS Paper Checklist}

\begin{enumerate}

\item {\bf Claims}
    \item[] Question: Do the main claims made in the abstract and introduction accurately reflect the paper's contributions and scope?
    \item[] Answer: \answerYes{}
    \item[] Justification: The abstract and Section~\ref{sec:intro} list three contributions: a reverse-probe contamination-audit pipeline, WirelessMathBench-XL as a wireless mathematical benchmark with per-problem contamination metadata, and reference WirelessMathLM release-split baselines. Section~\ref{sec:contamination-audit} and Section~\ref{sec:exp_contam} support the audit claim; Section~\ref{sec:dataset_construction} supports the benchmark claim; Section~\ref{sec:exp_main}--Section~\ref{sec:exp_transfer} support the WirelessMathLM rows as supporting evidence. Scope boundaries for the contamination claim, the verifier claim, and the release-split training claim are stated near the corresponding results in the main text and collected in Appendix~\ref{app:use_cases}.

\item {\bf Limitations}
    \item[] Question: Does the paper discuss the limitations of the work performed by the authors?
    \item[] Answer: \answerYes{}
    \item[] Justification: Section~\ref{sec:conclusion} and Appendix~\ref{app:use_cases} state: locked scores are defined under one semantic judge, with an early cross-judge probe ($3.4$--$9.3\%$ absolute worst case per model, Appendix~\ref{app:evaluation_robustness_app}); headline $\mathcal{S}_0/\mathcal{S}_1$ membership is defined against RedPajama-arXiv, with OpenWebMath as an off-domain corroboration run, so the audit can miss exposure in closed pretraining mixes; overlap is a proxy for memorisation, not a measurement of it; the retained expert-screening records have low retrospective agreement (binary inclusion $\kappa=0.164$, Appendix~\ref{app:expert_agreement}); the completed size-matched general-math GRPO control is scored on WMB-XL but not yet on the same five transfer benchmarks (Section~\ref{sec:exp_transfer}).

\item {\bf Theory assumptions and proofs}
    \item[] Question: For each theoretical result, does the paper provide the full set of assumptions and a complete (and correct) proof?
    \item[] Answer: \answerNA{}
    \item[] Justification: The paper presents an empirical dataset, an audit protocol, and trained reference models. No theorems or formal proofs are claimed.

\item {\bf Experimental result reproducibility}
    \item[] Question: Does the paper fully disclose all the information needed to reproduce the main experimental results of the paper to the extent that it affects the main claims and/or conclusions of the paper?
    \item[] Answer: \answerYes{}
    \item[] Justification: The construction pipeline is described in Section~\ref{sec:dataset_construction} with prompts in Appendix~\ref{app:prompt_construction}, the rubric in Appendix~\ref{app:quality_rubric_human}, and the system-model extraction examples in Appendix~\ref{app:model_examples}. The contamination audit is specified in Section~\ref{sec:contamination-audit} and Appendix~\ref{app:audit-method}, with corroboration and negative-control runs in Appendix~\ref{app:audit-iterations}. The GRPO recipe is in Section~\ref{sec:exp_setup} with full hyperparameters in Appendix~\ref{app:training_details}.

\item {\bf Open access to data and code}
    \item[] Question: Does the paper provide open access to the data and code, with sufficient instructions to faithfully reproduce the main experimental results?
    \item[] Answer: \answerYes{}
    \item[] Justification: The dataset is public on the Hugging Face Hub and the audit, split, and paired-bootstrap scripts and per-model result tables on GitHub (Appendix~\ref{app:datasheet}), with a Croissant~\citep{akhtar2024croissant} record. Following source-paper licenses, $1{,}542$ of $4{,}027$ records are distributed and $2{,}485$ are withheld and listed by identifier only; per-problem audit metadata, split assignments, and $\mathcal{S}_0$ membership by $n$ cover all problems. Full-test-split results use the authors' complete copy; the $310$-item public test subset and per-model results on it (Appendix~\ref{app:public_subset}) give the externally reproducible view. WirelessMathLM checkpoints are released on the Hugging Face Hub.

\item {\bf Experimental setting/details}
    \item[] Question: Does the paper specify all the training and test details necessary to understand the results?
    \item[] Answer: \answerYes{}
    \item[] Justification: Section~\ref{sec:exp_setup} reports model sets, evaluation protocol, GRPO recipe (composite reward, optimiser, learning rate, schedule, KL penalty, epochs / steps), generation temperature, and verifier configuration. Appendix~\ref{app:training_details} contains the full hyperparameter table. The data split is specified in Section~\ref{sec:stats}.

\item {\bf Experiment statistical significance}
    \item[] Question: Does the paper report error bars suitably and correctly defined or other appropriate information about the statistical significance of the experiments?
    \item[] Answer: \answerYes{}
    \item[] Justification: The contamination-aware re-evaluation (Section~\ref{sec:exp_contam}, Tabs.~\ref{tab:main_results} and~\ref{tab:wml7b_pairs}) reports paired bootstrap $95\%$ confidence intervals on per-model accuracy differences, computed from $10{,}000$ bootstrap resamples drawn at the problem level (resample with replacement; same indices applied to both models). These intervals capture sampling variation under the locked judge; the early cross-judge probe ($3.4$--$9.3\%$ absolute worst case per model, Appendix~\ref{app:evaluation_robustness_app}) is reported separately. Error bars on headline single-run accuracies are not reported because frontier/proprietary API and large hosted-model costs preclude multi-run replication within the available evaluation budget; the paired-bootstrap protocol of Section~\ref{sec:exp_contam} provides the statistical instrument the headline claims rely on. Training comparisons use one run per condition.

\item {\bf Experiments compute resources}
    \item[] Question: For each experiment, does the paper provide sufficient information on the computer resources needed to reproduce the experiments?
    \item[] Answer: \answerYes{}
    \item[] Justification: GRPO training compute is in Section~\ref{sec:exp_setup}: $4 \times$ NVIDIA RTX 4090 D, with $28.4$\,h / $53.4$\,h for $3$/$7$\,B. Evaluation hardware is in Appendix~\ref{app:cleaned_detail}: local open-weights inference uses $4 \times$ NVIDIA RTX 4090 D ($48$\,GB each), while closed-source and $\geq 70$\,B rows use vendor APIs. Contamination audit compute is in Section~\ref{sec:contamination-audit} / Appendix~\ref{app:audit-method}: one workstation with a $16$-worker implementation streams the $12.9$\,B RedPajama-arXiv $13$-grams at about $6$\,GB resident. The algebraic-stack negative-control run (Appendix~\ref{app:audit:algebraic-stack}) completed in $412$\,s and returned $2/4027$ hits ($0.05\%$); the OpenWebMath corroboration run returned $16/4027$ hits ($0.40\%$).

\item {\bf Code of ethics}
    \item[] Question: Does the research conducted in the paper conform, in every respect, with the NeurIPS Code of Ethics?
    \item[] Answer: \answerYes{}
    \item[] Justification: The dataset is derived from public arXiv preprints; no human-subjects data or personally identifiable information are collected. Expert validation was performed by project collaborators as research QA on mathematical artifacts, not by paid crowdworkers. The release contains derived mathematical problems and audit metadata rather than operational wireless procedures, exploit instructions, or controlled technical designs.

\item {\bf Broader impacts}
    \item[] Question: Does the paper discuss both potential positive societal impacts and negative societal impacts?
    \item[] Answer: \answerYes{}
    \item[] Justification: Positive: enabling contamination-aware evaluation in technical-domain LLM benchmarks (the audit pipeline is reusable beyond wireless), and providing a wireless-engineering-copilot reference (Appendix~\ref{app:use_cases}). Negative: a strong wireless-math model could in principle accelerate engineering automation that displaces practitioner labour; we judge this risk to be diffuse rather than acute given the technical scope.

\item {\bf Safeguards}
    \item[] Question: Does the paper describe safeguards that have been put in place for responsible release of data or models that have a high risk for misuse?
    \item[] Answer: \answerNA{}
    \item[] Justification: The dataset is mathematical content derived from public arXiv preprints; the model class (small-to-mid-sized math-specialised LLMs) does not pose a high-risk misuse profile (no frontier general-purpose model release, no scraped personal data, no image generation).

\item {\bf Licenses for existing assets}
    \item[] Question: Are the creators or original owners of assets used in the paper properly credited and are the license and terms of use explicitly mentioned and properly respected?
    \item[] Answer: \answerYes{}
    \item[] Justification: arXiv source papers are identified by the released \texttt{paper\_id} / arXiv-identifier fields and their text is not redistributed. We harvested each source paper's license from arXiv OAI-PMH and distribute a derived record only when that license clearly permits it (CC~BY, CC0, CC~BY-SA, CC~BY-NC-SA, each released under the matching license); records from arXiv-default, CC~BY-NC-ND, or unlabeled sources are withheld. Pretraining-slice references: RedPajama~\citep{weber2024redpajama}, Dolma~\citep{soldaini2024dolma}, ProofPile-2 / Llemma~\citep{azerbayev2024llemma}. Base models: Qwen2.5-0.5B, Qwen2.5-7B, and Qwen3-4B (Apache-2.0); Qwen2.5-3B (Qwen Research License), which also governs the released WirelessMathLM-3B checkpoint. Evaluator: GPT-4.1-mini via OpenAI API (subject to OpenAI's API ToS). Details are in Appendix~\ref{app:datasheet} (Distribution section).

\item {\bf New assets}
    \item[] Question: Are new assets introduced in the paper well documented and is the documentation provided alongside the assets?
    \item[] Answer: \answerYes{}
    \item[] Justification: The dataset ships with a Datasheet for Datasets record~\citep{gebru2021datasheets} (Appendix~\ref{app:datasheet}), a Croissant~\citep{akhtar2024croissant} metadata record, per-problem contamination metadata (Section~\ref{sec:contamination-audit}), audit and evaluation scripts with per-model result tables, and WirelessMathLM training hyperparameters (Appendix~\ref{app:training_details}). Distributed records carry per-record source and record licenses (CC~BY~4.0, CC~BY-SA~4.0, or CC~BY-NC-SA~4.0 matching the source paper); identifier-level metadata is CC~BY~4.0 and code is MIT; source-paper text is not redistributed (Appendix~\ref{app:datasheet}, Distribution section).

\item {\bf Crowdsourcing and research with human subjects}
    \item[] Question: For crowdsourcing experiments and research with human subjects, does the paper include the full text of instructions given to participants and screenshots, if applicable, as well as details about compensation?
    \item[] Answer: \answerNA{}
    \item[] Justification: No crowdsourcing and no human-subjects data collection. The Tier-2 expert review is conducted by five PhD-student or postdoctoral domain specialists (Section~\ref{sec:qa}) as research QA on mathematical artifacts, not as a crowdworker study. The rubric and instructions they apply are in Appendix~\ref{app:quality_rubric_human}; coverage and agreement analysis is in Appendix~\ref{app:expert_agreement}.

\item {\bf Institutional review board (IRB) approvals or equivalent for research with human subjects}
    \item[] Question: Does the paper describe potential risks incurred by study participants, whether such risks were disclosed to the subjects, and whether IRB approvals were obtained?
    \item[] Answer: \answerNA{}
    \item[] Justification: No human-subjects research is conducted: the work annotates mathematical artifacts and uses expert validation by project collaborators, not study participants. No personal data are collected.

\item {\bf Declaration of LLM usage}
    \item[] Question: Does the paper describe the usage of LLMs if it is an important, original, or non-standard component of the core methods?
    \item[] Answer: \answerYes{}
    \item[] Justification: LLMs are used in three core-method roles, all declared. (i) Stage-two source-paper filter: GPT-4o (Section~\ref{sec:collection}). (ii) Structured model extraction: DeepSeek-R1~\citep{guo2025deepseek} (Section~\ref{sec:extraction}). (iii) Tier-1 rubric scoring and evaluation judge: GPT-4o for Tier-1, GPT-4.1-mini for evaluation-time judge (Section~\ref{sec:qa}, Section~\ref{sec:exp_setup}; cross-evaluator robustness to Claude-3-Haiku and Gemini-2.0-Flash in Appendix~\ref{app:evaluation_robustness_app}). LLM use for writing / editing the manuscript itself is governed by the standard NeurIPS LLM policy and is documented in Appendix~\ref{app:llm_disclosure}.

\end{enumerate}

\end{document}